\documentclass[journal,twoside]{IEEEtran}
\usepackage[utf8]{inputenc}
\usepackage[T1]{fontenc}

\usepackage{graphicx} %
\usepackage{tabularx}
\usepackage{hyperref}
\usepackage{times} %
\usepackage{amsmath} %
\usepackage{amssymb}  %
\usepackage[mathcal]{eucal}
\usepackage{bm}
\usepackage{multirow}
\usepackage{xspace}
\usepackage{stfloats}
\usepackage{url}
\usepackage[table,dvipsnames]{xcolor}
\usepackage{xcolor}
\usepackage{amsfonts}
\usepackage{import}
\usepackage{siunitx}
\usepackage{soul}
\usepackage{ulem}
\usepackage{xargs}

\hypersetup{
    colorlinks,
    linkcolor={black!50!black},
    citecolor={blue!50!black},
    urlcolor={blue!80!black}
}

\newcommand{\bs}[1]{\boldsymbol{#1}}
\newcommand{\q}{\bm{q}}
\newcommand{\dq}{\dot{\q}}

\newcommand{\M}{\bm{M}}

\newcommand{\K}{\bm{K}}
\newcommand{\tauv}{\bm{\tau}}
\newcommand{\thm}{\bm{\theta}}
\newcommand{\dthm}{\dot{\thm}}
\newcommand{\ddthm}{\ddot{\thm}}

\newcommand{\Jac}{\bm{J}}

\newcommand{\parX}[2]{\frac{\partial #1}{\partial #2}}
\newcommand{\half}{\frac{1}{2}}

\newcommand{\tr}{\top}
\newcommand{\ii}{\mathrm{i}}

\newcommand{\inR}[1]{\in{\mathbb{R}}^{#1}}
\newcommand{\cc}{\bm{c}}
\newcommand{\gv}{\bm{g}}

\newcommand{\vv}{\bm{v}}

\newcommand{\xeq}{\x_\mathrm{eq}}

\newcommand{\xturn}{\q_{\curvearrowleft}}
\newcommand{\GG}{\mathcal{G}}
\newcommand{\x}{\bm{x}}
\newcommand{\dx}{\dot{\x}}
\newcommand{\ddx}{\ddot{\x}}
\newcommand{\z}{\bm{z}}
\newcommand{\dz}{\dot{\z}}
\newcommand{\ddz}{\ddot{\z}}
\newcommand{\ebert}{\textit{eBert}\xspace}

\title{Exploring Nonlinear Body Oscillations\\for Natural Quadruped Gaits}

\author{
	Annika Schmidt${}^{1,2,\ast}$\thanks{${}^{\ast}$Corresponding author: \href{mailto:annika.schmidt@dlr.de}{annika.schmidt@dlr.de}}\thanks{${}^1$Technical University of Munich (TUM), Department of Computer Engineering, Germany.}\thanks{${}^{2}$German Aerospace Center (DLR), Institute of Robotics and Mechatronics, Weßling, Germany.},
	Davide Calzolari${}^{1,2}$,
	Arne Sachtler${}^{1,2}$,
	Florian Loeffl${}^{2}$,
	Daniel Seidel${}^{2}$,
	Milan Hermann${}^{2}$,
	{Robert Burger}${}^{2}$,
	{Thomas Gumpert}${}^{2}$,
	{Antonin Raffin}${}^{2}$,
	{Tristan Ehlert}${}^{2}$,
	{Maximilian Pries}${}^{2}$,
	{David Wandinger}${}^{2}$,
	{Florian Schmidt}${}^{2}$,
	Manuel Keppler${}^{2,3}$\thanks{${}^{3}$Eindhoven University of Technology (TU/e), Department of Mechanical Engineering, The Netherlands.},
	Jinoh Lee${}^{2.4}$\thanks{${}^{4}$Korea Advanced Institute of Science \& Technology (KAIST), Department of Mechanical Engineering, South Korea.}, and
	Alin Albu-Schäffer${}^{1,2}$%
}

\begin{document}

\maketitle
\thispagestyle{empty}

\markboth{Extended Preprint}{Schmidt \MakeLowercase{\textit{et al.}}: Exploring Nonlinear Body Oscillations for Natural Quadruped Gaits}

{\renewcommand{\abstractname}{Summary}
\begin{abstract}
The research demonstrates how the analysis of nonlinear normal modes aids to develop natural, multi-gait locomotion in a compliant quadruped robot leveraging its inherent mechanics, offering a design tool to shape embodied intelligence in robotics.
\end{abstract}}

\begin{abstract}
Animals’ body morphology shapes the gait patterns they can perform, where mechanical resonance reduces the need for active control. By tuning posture and muscle stiffness, they leverage their embodied intelligence to achieve effective gaits for different speeds. In contrast, most quadruped robots are not specifically designed to exploit mechanical resonance due to the complexity of nonlinear dynamics and require dedicated locomotion controllers. To provide an alternative, we present a proof-of-concept framework making the nonlinear dynamics of a robot predictable in the design process and show how this knowledge can be leveraged such that multi-gait locomotion can emerge from nonlinear resonances—shaped by gravity, inertia, and elasticity.
We present the highly compliant quadruped robot \ebert, on which we identify six nonlinear normal modes (NNMs) using our new theoretical tools and validate their existence in simulation and hardware.
With black-box optimization to determine step length, simulations show how each NNM naturally develops into a distinct gait, manifesting different speeds, which also largely transfers to the robotic hardware. Our experiments show that \ebert can exploit its mechanics to generate task-specific movements which may serve as foundation for designing a new generation of agile and efficient robots leveraging embodied intelligence.
\end{abstract}

\section{Introduction}
\IEEEPARstart{Q}{uadrupedal} robots hold great potential for real-world applications due to their inherent stability and weight-bearing capabilities, leading to a rapidly expanding market \cite{bis21, chi23}. 
Despite improving capabilities, %
most quadrupeds converge on similar designs driven by stiff actuation with high-gain feedback, yielding gaits dictated by the controller rather than mechanics \cite{fuk22}.
In contrast, animals achieve efficient, habitat-specific movement \cite{zef03} through tight coupling between morphology and neural wiring \cite{niven2008, valero2022bio}.
This integration enables locomotion to emerge from the natural interplay of mechanics and control \cite{pfeifer2007self} - a principle known as \textit{embodied intelligence} \cite{Brooks1991aiBody}.
One key aspect is that an animal's morphology not only supports locomotion but also constrains and shapes its dynamics. Thus, an animal's body structure determines which gaits it can perform \cite{blickhan2021trunk} and which are preferred at different speeds \cite{van01, kur08}.
Hoyt and Taylor famously demonstrated this in horses, showing that switching gaits at specific speeds minimized energy costs \cite{hoy81}.
To maintain energy efficiency across speeds, animals adjust their posture, gait patterns, and muscle stiffness, effectively tuning their body’s resonance \cite{alexander2003principles, Cavagna2015running}, while the elasticity in tendons and muscles enhances robustness by absorbing impact energy \cite{alexander1988elastic}.

\begin{figure*}
    \centering
    \includegraphics[width=\textwidth]{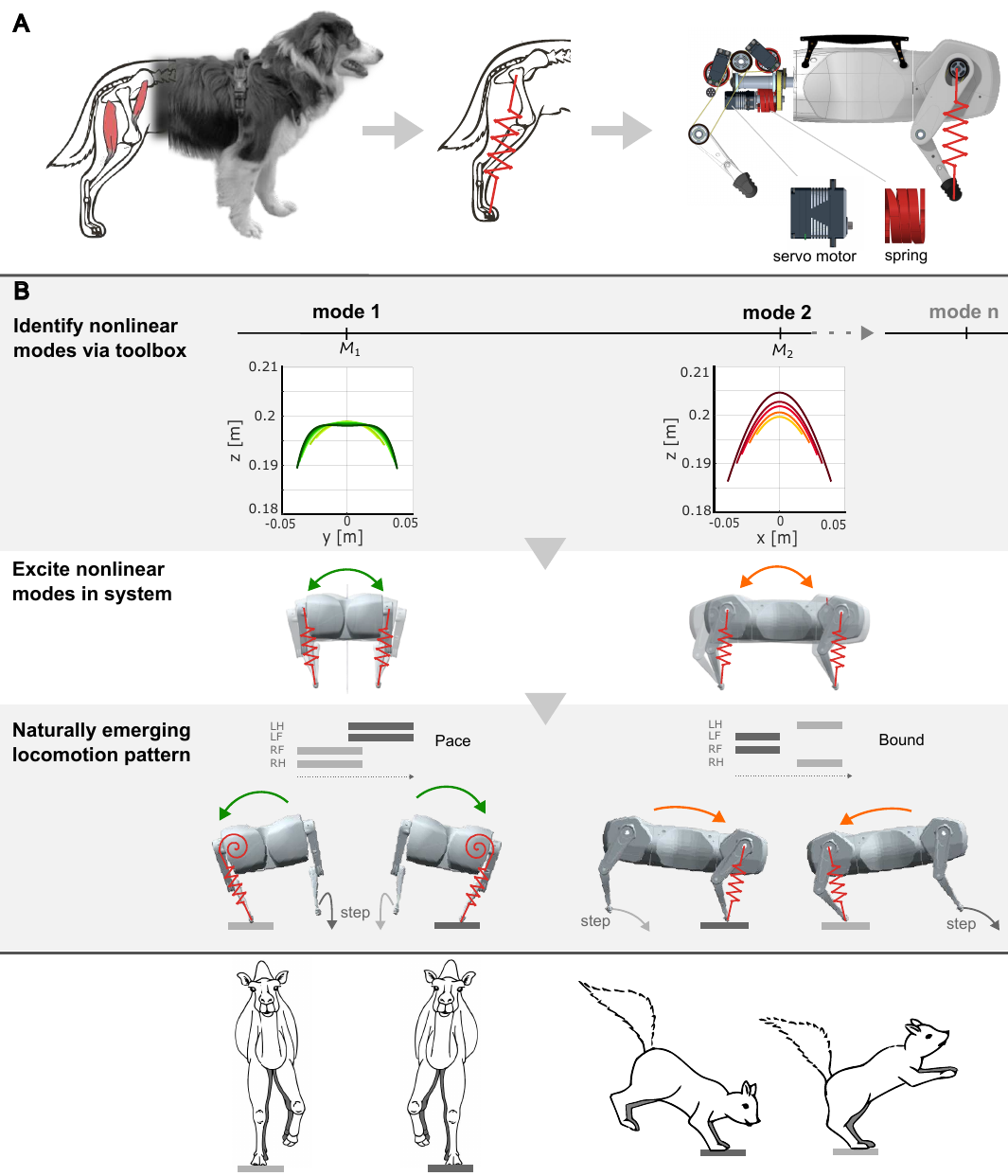}
    \caption{\textbf{Workflow to develop gaits based on embodied intelligence in \ebert.} \textbf{(A)} Inspired by the elasticity of
    biological tendons and muscles, the quadruped robot \ebert features series elastic actuation with soft springs to implement this function mechanically.
    High compliance in the hardware design results in a system that naturally displays large amplitude oscillations in different directions.
    \textbf{(B)} Leveraging the biological concept of embodied intelligence, the natural oscillations encoded in the robotic hardware can be exploited following three key aspects:
    First, the mathematical description and analysis of the system motions; second, the excitation of nonlinear resonance; and finally, the generation of (loco)motion patterns, leading to gaits that resemble the ones known in animals.\\ }
    \label{fig:bioinsp-bert}
\end{figure*}

Motivated by these biomechanical insights, a line of research on legged locomotion has focused on including elastic elements in robotic designs \cite{del20, zha20, bad22, frund2023bipedal} and leveraging the added compliance through matched control strategies \cite{rut08, che19, cal23, ijs23}. This extends early work on passive dynamic walkers that exploit intrinsic mechanical responses to generate inherently stable, energy-efficient locomotion, with dynamics well approximated by an inverted pendulum \cite{McGeer1990}. Adding a spring to the simple walking template resulted in the \textit{spring-loaded inverted pendulum} (SLIP) model \cite{Blickh1989, poulakakis2009spring, Geyer2006} enabling the characterization of multiple gaits and their transitions.
Several robotic experiments validated that exploiting elasticity increases energy efficiency and robustness \cite{bad22, Hutter2012, ruppert2022nature, stella2025synergy}. %
Simulations of an elastic sagittal quadruped with prismatic legs showed that minimizing energy consumption with optimal control naturally produces different gaits at varying speeds \cite{remy16, raff2022generating}, consistent with biological observations.
Adaptive oscillators, similar to the biological principle of \textit{Central Pattern Generators} (CPG), proved effective to extract and excite the intrinsic frequencies of robotic hardware \cite{ijspeert2008central, ram23}.
Notable examples include the works of Badri-Spröwitz and Ijspeert demonstrating that locomoting robots inspired by biological mechanics can display natural gaits with remarkably simple CPG controllers \cite{bad22, ruppert2022nature, ijspeert2008central}. Well-designed systems such as BirdBot can exhibit gaits even under open-loop control \cite{bad22}, but feedback remains generally essential to entrain control frequency to intrinsic system dynamics and exploit compliance, as demonstrated on the quadruped Morti \cite{ruppert2022nature}.
However, without explicit engineering, compliant hardware typically converges to a single gait pattern dictated by its dominant mechanical resonance.

Achieving different gaits through hardware properties requires meticulous design derived from detailed analysis %
of biological principles, as exemplified by the synergies implemented in the quadruped robot PAWS \cite{stella2025synergy}.
More generally, designing legged robots where the compliance enables multiple gaits remains a challenge \cite{fuk22} because its typically nonlinear dynamics are difficult to predict in advance.

Recent advances in nonlinear mode theory enabled us to address this challenge and design hardware with diverse purposeful dynamics. Using mathematical tools based on differential geometry and algebraic topology \cite{alb20, sac22}, we can analyze nonlinear oscillations in complex systems \cite{bje22} to identify families of periodic orbits in energy-conservative systems, including Nonlinear Normal Modes (NNMs) \cite{alb21}, which generalize linear eigenmodes to nonlinear dynamics.
This offers a framework to systematically compute the natural dynamics even in complex, high-dimensional systems with significant nonlinearities. Building on the biological insight that efficient locomotion is tightly connected to exploiting body resonance \cite{van01, alexander2003principles}, we hypothesize that common quadrupedal gaits~--~such as trotting, pacing, and bounding~--~may naturally arise from NNMs in purposefully designed quadrupeds.
While the theoretical foundation of the NNM analysis has been established in prior work \cite{alb20, bje22}, this paper presents the first application of the complete framework to a full 3D quadruped robot to develop locomotion. %
This includes:
A) the design of the highly elastic quadruped \ebert;
B) the analysis of \ebert's nonlinear modes uncovering the robot's natural movement tendencies; and
C) the excitation of these modes to generate gaits using only minimal control.

Starting from biological leg stiffness approximations, \ebert's design encodes functional oscillations through its mechanics (Fig.~\ref{fig:bioinsp-bert}A). Using our developed methods~\cite{bje22, alb21, alb20}, we identify six NNMs of the conservative dynamics, which we validate in simulations and hardware, excited by a deliberately simple feedback-based state-switching controller to compensate for friction.
Extending this controller to lift the feet according to a pattern matching each mode enables distinct, and biologically plausible gaits to emerge naturally from the robot’s dynamics (Fig.~\ref{fig:bioinsp-bert}B).

With our hypothesis successfully validated, \ebert is, to the best of our knowledge, the first full robotic quadruped demonstrating that diverse gaits naturally emerge from purposefully designed nonlinear resonances, manifesting embodied intelligence in robotic hardware.
Rather than replicating animal locomotion in stride, speed, or efficiency, we claim to show that \ebert's emerging gaits follow biological principles where body resonance naturally organizes multi-legged coordination without explicit scripting.
Our framework can guide robotic co-design with intrinsic oscillations tailored to desired tasks, while improved dynamics analysis can deepen understanding of biological movement, advancing both robotic and biological knowledge.

\section{Results}
\subsection{The highly elastic quadruped \ebert}

Animal gaits can be viewed as periodic oscillations that exploit resonance of the musculoskeletal system to minimize energy expenditure \cite{alexander2003principles, Cavagna2015running}.
Similarly, compliant robots can improve efficiency by exciting their natural resonances with controllers entrained to the mechanics
\cite{ijspeert2008central,raffin2023simple}, but usually only one or few gaits arise from the most prominent resonance \cite{bad22, ruppert2022nature}.
A richer locomotion repertoire based on resonances requires a robot that mechanically encodes multiple useful high-amplitude body oscillations, which we realized with the highly-compliant quadruped \ebert with 12 degrees of freedom (DOF).

\begin{figure*}
    \centering
    \includegraphics[width=\textwidth]{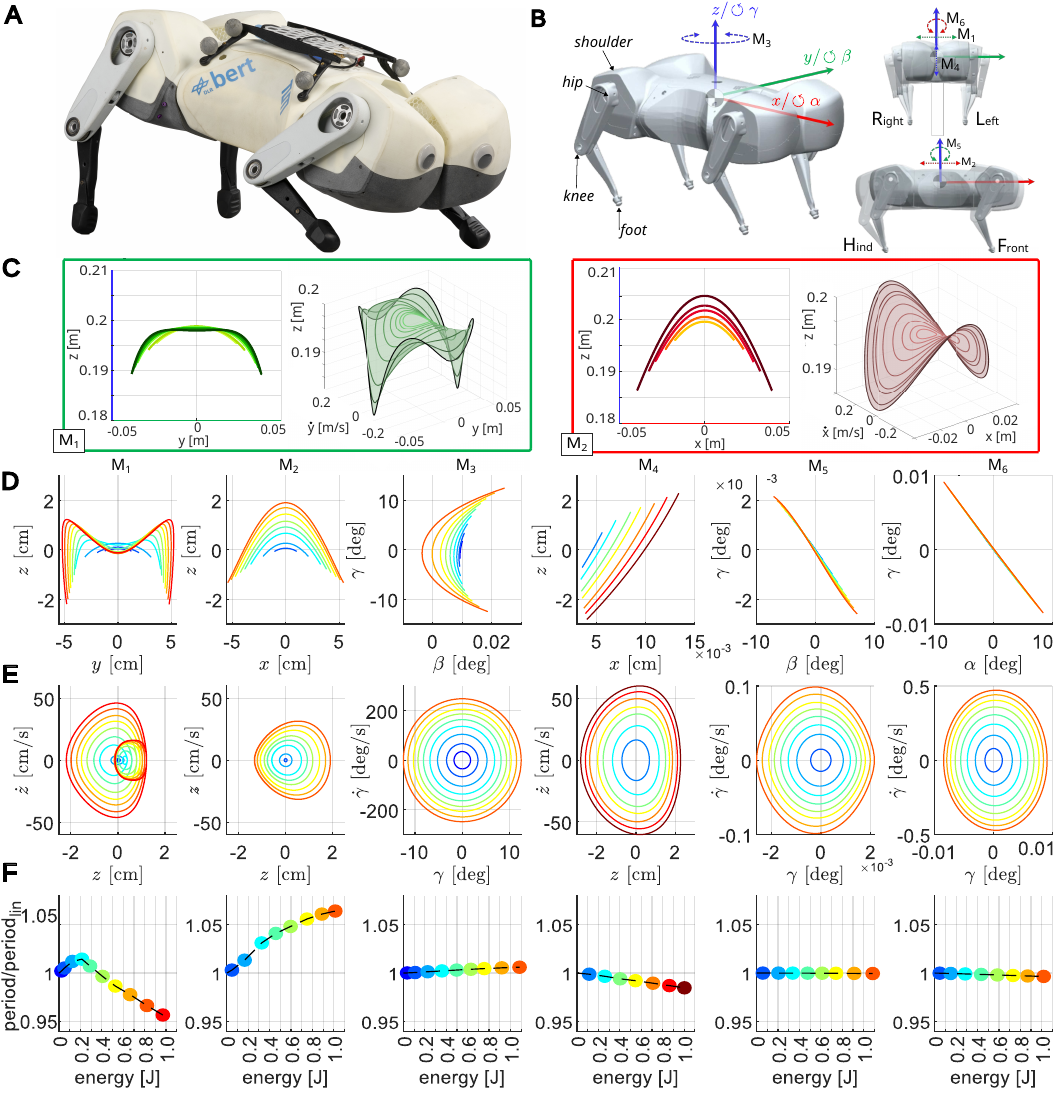}
    \caption{
    \textbf{Overview of \ebert and the nonlinear normal modes encoded in its mechanics.}
    \textbf{(A)} Hardware of the highly elastic quadruped robot \ebert crouching.
    \textbf{(B)} Indication of axes and motion directions of \ebert with modes $\mathcal{M}_1$-$\mathcal{M}_6$ from different perspectives.
    \textbf{(C)} Exemplary depiction of mode oscillations of $\mathcal{M}_1$ (left) and $\mathcal{M}_2$ (right), each orbit computed as solution to the equations of motion of the energy-conservative dynamics requiring no control.
    Plotting motions over respective relevant axes and in state space projections visualizes the nonlinearity.
     \textbf{(D)} Visualization of each mode by plotting the motion of the robot's center of mass in the respective relevant axes for different energy levels apparent from (F).
     \textbf{(E)} Phase diagram of each mode plotted in the relevant axes along which each mode oscillation develops with increasing energy.
     \textbf{(F)} Progression of the period of each mode over different energy levels. The y-axis displays the change of period relative to the period determined for the linearized system ($y=1.00$).
    }
    \label{fig:bert-modes}
\end{figure*}

\ebert was designed with lightweight elastic legs and large enough compliance to encode multiple functional body oscillations at frequencies and amplitudes suitable for gait generation (Fig.~\ref{fig:bert-modes}A,B).
The precise leg stiffness of \ebert was estimated assuming the body dimensions of a small dog weighing \SI{4.5}{\kilo \gram} using calculations for animal leg stiffness \cite{farley1993jeb}. Adapting the initial stiffness estimation such that \ebert could stand upright without active corrections led to a stiffness of \SI{6.55}{\newton \meter \per \radian} in the \textit{hip} and \textit{knee} joints, and \SI{13}{\newton \meter \per \radian} in the \textit{shoulders}. Design and stiffness details are in the Method section \textit{\nameref{sec:method:stiffness}}.
The hardware elasticity was implemented using Serial Elastic Actuators (SEAs) with springs having the respective stiffness at each joint. This can mimic the energy storage of animals' muscles and tendons but, unlike biology, SEA stiffness cannot be adjusted and is joint-local instead of distributed throughout the body \cite{alexander2003principles}.
Nevertheless, SEAs provide a practical way to add mechanical compliance in robots \cite{kim2013soft} (Fig.~\ref{fig:bioinsp-bert}A).
With the implemented compliance \ebert encodes large mechanical oscillations in various directions, yielding much higher system compliance than previous elastic quadrupeds exploiting embodied intelligence \cite{ruppert2022nature, stella2025synergy}.

Replicating embodiment in robotics also requires control that matches the system dynamics.
While in animal bodies mechanics and neural control are tightly coupled, robotics allows separate analysis of each to aid better understanding of individual contributions.
In \ebert we focused on the mechanical aspects encoding oscillations through design, only applying minimalistic control. We first analyzed the inherently nonlinear system dynamics of \ebert enabled by our recently developed numerical methods \cite{bje22, alb20} to identify the system's \textit{Nonlinear Normal Modes} (NNMs).
These NNMs extend the concept of linear normal modes, which characterize periodic coordinated oscillations of a mechanical system around an equilibrium with a constant amplitude and phase.
In contrast, NNMs vary with energy levels changing shape and period on curved surfaces called \textit{Eigenmanifolds} \cite{alb20} (Fig.~\ref{fig:bert-modes}C).
Using our methods, the NNMs of the conservative model can be computed, which describe intrinsic oscillations purely from the robot’s mechanical properties, without control input.
Computational details are outlined in the Supplementary Methods \textit{\nameref{sec:app:nnm}}.

We hypothesized that distinct gaits can emerge from the different NNMs.
As starting point, we analyzed \ebert's natural motion tendencies around a symmetric standing position (shoulders neutral, hips and knees slightly flexed, Fig.~\ref{fig:bert-modes}B) with the commanded motor positions $\bm{\theta}_0$ for the three joints per leg being
\begin{equation}
\begin{aligned}
    \bm{\theta}_{0, \mathrm{FR}} &= \bm{\theta}_{0,\mathrm{FL}} = \bm{\theta}_{0,\mathrm{HR}} = \bm{\theta}_{0,\mathrm{HL}} \\
    &= [ \underset{\substack{\uparrow\\\mathrm{shoulder}}}{0,} \ \ \underset{\substack{\uparrow\\\mathrm{hip}}}{+0.4,}  \ \ \underset{\substack{\uparrow\\\mathrm{knee}}}{-0.4 } \ ]^\tr \ \mathrm{rad} \\
    &\approx [ \underset{\substack{\uparrow\\\mathrm{shoulder}}}{0,} \ \ \underset{\substack{\uparrow\\\mathrm{hip}}}{+23^{\circ},}  \ \ \underset{\substack{\uparrow\\\mathrm{knee}}}{-23^{\circ}} \ ]^\tr \ ,
\end{aligned}
\label{eq:leg_angle_def}
\end{equation}
where subscripts $\mathrm{F}$ and $\mathrm{H}$ refer to the robot's front and hind leg, respectively, while $\mathrm{L}$ and $\mathrm{R}$ denote the left and right sides, relative to the motion direction.
Assuming fixed foot positions but free rotations yields a constrained system with $n = 6$~DOFs and six corresponding linear modes, from which at least six NNMs were expected based on the Seifert conjecture \cite{alb20}.

The six computed NNMs are displayed in Fig.~\ref{fig:bert-modes}, with the associated eigenvectors $w$ and their characteristic oscillation frequency of the linearized system summarized on the left side of Table~\ref{tab:linmodes-freq}.
As predicted by theory, NNM frequency and shape vary with energy as the system enters the nonlinear regime at larger deflections (Fig.~\ref{fig:bert-modes}D,F).
Sorting the NNMs by frequency revealed that the slowest mode ($\mathcal{M}_1$) is a translation of the robot body along the y-axis coupled with rolling (Fig.~\ref{fig:bert-modes}B,D).
The second mode ($\mathcal{M}_2$) couples a pitching oscillation with a forward-backward motion along the x-axis.
The third mode ($\mathcal{M}_3$) manifests as rotation around the z-axis going vertically through the body's center of mass (COM). The fourth motion ($\mathcal{M}_4$) is a translational up-and-down motion along the z-axis. The fifth ($\mathcal{M}_5$) and sixth modes ($\mathcal{M}_6$) are the pure pitch and roll motions, respectively, both oscillating at much higher frequencies (Tab.~\ref{tab:linmodes-freq}, left bottom).
Especially the first two modes show large bending for higher energy levels, while the remaining modes only display small displacements in Cartesian space due to the higher oscillation frequencies (Fig.~\ref{fig:bert-modes}D).
The phase plot of each NNM for the respective dominant motion direction (Fig.~\ref{fig:bert-modes}E) still verifies the expected nonlinearity for all modes except  $\mathcal{M}_6$,
which appears to adhere to the special case of a \textit{strict mode}~\cite{alb21,sac22}, which in this case also resembles a linear mode (Fig.~\ref{fig:bert-modes}D) and orbits project to concentric ellipses in the phase plot (Fig.~\ref{fig:bert-modes}E).
The Supplementary Methods \textit{\nameref{sec:app:quantify-nonlinearity}} present more detailed quantification of the modes' nonlinearity.

In essence, we demonstrated that \ebert's design encodes multiple distinct NNMs with large enough amplitudes, expected to serve as foundations for different gaits.
\begin{table*}
\centering
\caption{\textbf{Descriptive values of modal oscillations of \ebert}. Values are presented for the initial standing pose (left) and the configuration used for stepping forward (right) calculated for the conservative model, in simulation with friction (light gray) and from the hardware of \ebert (dark gray).}
\label{tab:linmodes-freq}
\begin{tabular}{l|cccccrl|clclcl}
                                       & \multicolumn{6}{c}{\textbf{Standing Leg Config}}                                                                                                                  & \multicolumn{1}{c}{\textbf{}} & \multicolumn{6}{|c}{\textbf{Stepping Leg Config}}                                                                                                                                          \\
                                    & \multicolumn{6}{c}{$\bm{\theta}_{0,F} = [\ 0.0, \ 0.4, -0.4]^\tr$ }                                                                                                                  & \multicolumn{1}{c}{\textbf{}} & \multicolumn{6}{|c}{$\bm{\theta}_{0,F} = [-0.1, \ 0.4, -0.4]^\tr$ }                                                                                                                                          \\
                                    & \multicolumn{6}{c}{$\bm{\theta}_{0,H} = [\ 0.0, \ 0.4, -0.4]^\tr$ }                                                                                                                  & \multicolumn{1}{c}{\textbf{}} & \multicolumn{6}{|c}{$\bm{\theta}_{0,H} = [-0.1, \ 0.4, -0.3]^\tr$ }                                                                                                                                          \\
                                          & \multicolumn{6}{c}{\includegraphics[height=2.2cm]{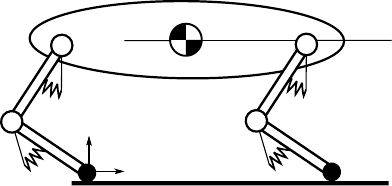}}                                                  & \multicolumn{1}{c}{\textbf{}} & \multicolumn{6}{|c}{\includegraphics[height=2.7cm]{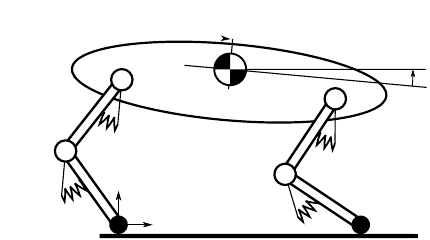}}                  \\[8pt]
\textbf{Mode}                          & \multicolumn{6}{c}{\textbf{Osc. Frequency in place {[\si{\hertz}]}}}                                                                                                & \multicolumn{1}{c}{\textbf{}} & \multicolumn{6}{|c}{\textbf{Osc. Frequency in place {[\si{\hertz}]}}}                                                                                                                         \\
                                       & \multicolumn{2}{c}{\cellcolor[HTML]{FFFFFF}conserv. model}                          & \multicolumn{2}{c}{\cellcolor[HTML]{E0E0E0}simulation}  & \multicolumn{2}{c}{\cellcolor[HTML]{D0D0D0}hardware}       &                               & \multicolumn{2}{c}{\cellcolor[HTML]{FFFFFF}conserv. model}                                                & \multicolumn{2}{c}{\cellcolor[HTML]{E0E0E0}simulation}      & \multicolumn{2}{c}{\cellcolor[HTML]{D0D0D0}hardware}           \\
$\mathcal{M}_1 \ (y_{\mathrm{trans}})$ & \multicolumn{2}{c}{\cellcolor[HTML]{FFFFFF}0.99}                           & \multicolumn{2}{c}{\cellcolor[HTML]{E0E0E0}0.99} & \multicolumn{2}{c}{\cellcolor[HTML]{D0D0D0}0.8}      &                               & \multicolumn{2}{c}{\cellcolor[HTML]{FFFFFF}1.05}                                                & \multicolumn{2}{c}{\cellcolor[HTML]{E0E0E0}1.02}     & \multicolumn{2}{c}{\cellcolor[HTML]{D0D0D0}1.07}         \\
$\mathcal{M}_2 \ (x_{\mathrm{trans}})$ & \multicolumn{2}{c}{\cellcolor[HTML]{FFFFFF}1.56}                           & \multicolumn{2}{c}{\cellcolor[HTML]{E0E0E0}1.59} & \multicolumn{2}{c}{\cellcolor[HTML]{D0D0D0}1.5}      &                               & \multicolumn{2}{c}{\cellcolor[HTML]{FFFFFF}1.56}                                                & \multicolumn{2}{c}{\cellcolor[HTML]{E0E0E0}1.57}     & \multicolumn{2}{c}{\cellcolor[HTML]{D0D0D0}1.52}         \\
$\mathcal{M}_3 \ (z_{\mathrm{rot}})$   & \multicolumn{2}{c}{\cellcolor[HTML]{FFFFFF}3.22}                           & \multicolumn{2}{c}{\cellcolor[HTML]{E0E0E0}3.20} & \multicolumn{2}{c}{\cellcolor[HTML]{D0D0D0}2.50}     &                               & \multicolumn{2}{c}{\cellcolor[HTML]{FFFFFF}3.12}                                                & \multicolumn{2}{c}{\cellcolor[HTML]{E0E0E0}3.12}     & \multicolumn{2}{c}{\cellcolor[HTML]{D0D0D0}2.45}         \\
$\mathcal{M}_4 \ (z_{\mathrm{trans}})$ & \multicolumn{2}{c}{\cellcolor[HTML]{FFFFFF}3.70}                           & \multicolumn{2}{c}{\cellcolor[HTML]{E0E0E0}3.99} & \multicolumn{2}{c}{\cellcolor[HTML]{D0D0D0}3.60}     &                               & \multicolumn{2}{c}{\cellcolor[HTML]{FFFFFF}4.00}                                                & \multicolumn{2}{c}{\cellcolor[HTML]{E0E0E0}3.39}     & \multicolumn{2}{c}{\cellcolor[HTML]{D0D0D0}4.42}         \\
$\mathcal{M}_5 \ (y_{\mathrm{rot}})$   & \multicolumn{2}{c}{\cellcolor[HTML]{FFFFFF}5.88}                           & \multicolumn{2}{c}{\cellcolor[HTML]{E0E0E0}6.09} & \multicolumn{2}{c}{\cellcolor[HTML]{D0D0D0}5.30}     &                               & \multicolumn{2}{c}{\cellcolor[HTML]{FFFFFF}5.88}                                                & \multicolumn{2}{c}{\cellcolor[HTML]{E0E0E0}7.01}     & \multicolumn{2}{c}{\cellcolor[HTML]{D0D0D0}5.76}         \\
$\mathcal{M}_6 \ (x_{\mathrm{rot}})$   & \multicolumn{2}{c}{\cellcolor[HTML]{FFFFFF}8.33}                           & \multicolumn{2}{c}{\cellcolor[HTML]{E0E0E0}8.79} & \multicolumn{2}{c}{\cellcolor[HTML]{D0D0D0}6.60}     &                               & \multicolumn{2}{c}{\cellcolor[HTML]{FFFFFF}8.33}                                                & \multicolumn{2}{c}{\cellcolor[HTML]{E0E0E0}10.57}    & \multicolumn{2}{c}{\cellcolor[HTML]{D0D0D0}7.21}         \\
                                       & \multicolumn{1}{l}{}        & \multicolumn{1}{l}{} & \multicolumn{1}{l}{}    & \multicolumn{1}{l}{}   & \multicolumn{1}{l}{}       & \multicolumn{1}{l}{}    &                               & \multicolumn{1}{l}{}                       &                        & \multicolumn{1}{l}{}                 &               & \multicolumn{1}{l}{}                  &                  \\
                                       & \multicolumn{6}{c}{\textbf{Spring Exploitation} $\eta$ }                                                      & \textbf{}                     & \multicolumn{6}{c}{\cellcolor[HTML]{E0E0E0}\textbf{Stepping in Simulation}}                                                                                                                                          \\
                                       & \multicolumn{2}{c}{}                               & \multicolumn{2}{c}{\cellcolor[HTML]{E0E0E0}simulation}  & \multicolumn{2}{c}{\cellcolor[HTML]{D0D0D0}hardware}       &                               & \multicolumn{2}{c}{\cellcolor[HTML]{E0E0E0}Forward Velocity [\si{\meter \per \second}]} & \multicolumn{4}{c}{\cellcolor[HTML]{E0E0E0}Frequency [\si{\hertz}]}                                                    \\
                                       & \multicolumn{2}{c}{}                               & \multicolumn{2}{c}{\cellcolor[HTML]{E0E0E0}}     & \multicolumn{2}{c}{\cellcolor[HTML]{D0D0D0}}         &                               & \multicolumn{2}{c}{\cellcolor[HTML]{E0E0E0}}                        & \multicolumn{2}{c}{\cellcolor[HTML]{E0E0E0}dominant} & \multicolumn{2}{c}{\cellcolor[HTML]{E0E0E0}(underlying)} \\
$\mathcal{M}_1 \ (y_{\mathrm{trans}})$ & \multicolumn{2}{c}{}                               & \multicolumn{2}{c}{\cellcolor[HTML]{E0E0E0}0.86} & \multicolumn{2}{c}{\cellcolor[HTML]{D0D0D0}0.65} &                               & \multicolumn{2}{c}{\cellcolor[HTML]{E0E0E0}0.054}                   & \multicolumn{2}{c}{\cellcolor[HTML]{E0E0E0}4.13}     & \multicolumn{2}{c}{\cellcolor[HTML]{E0E0E0}(1.35)}       \\
$\mathcal{M}_2 \ (x_{\mathrm{trans}})$ & \multicolumn{2}{c}{}                               & \multicolumn{2}{c}{\cellcolor[HTML]{E0E0E0}0.79} & \multicolumn{2}{c}{\cellcolor[HTML]{D0D0D0}0.75}     &                               & \multicolumn{2}{c}{\cellcolor[HTML]{E0E0E0}0.098}                   & \multicolumn{2}{c}{\cellcolor[HTML]{E0E0E0}3.54}     & \multicolumn{2}{c}{\cellcolor[HTML]{E0E0E0}(1.74)}       \\
$\mathcal{M}_3 \ (z_{\mathrm{rot}})$   & \multicolumn{2}{c}{}                               & \multicolumn{2}{c}{\cellcolor[HTML]{E0E0E0}0.48} & \multicolumn{2}{c}{\cellcolor[HTML]{D0D0D0}0.50}     &                               & \multicolumn{2}{c}{\cellcolor[HTML]{E0E0E0}0.11}                    & \multicolumn{2}{c}{\cellcolor[HTML]{E0E0E0}2.90}     & \multicolumn{2}{c}{\cellcolor[HTML]{E0E0E0}}             \\
$\mathcal{M}_4 \ (z_{\mathrm{trans}})$ & \multicolumn{2}{c}{}                               & \multicolumn{2}{c}{\cellcolor[HTML]{E0E0E0}0.49} & \multicolumn{2}{c}{\cellcolor[HTML]{D0D0D0}0.81}     &                               & \multicolumn{2}{c}{\cellcolor[HTML]{E0E0E0}0.15}                    & \multicolumn{2}{c}{\cellcolor[HTML]{E0E0E0}3.60}     & \multicolumn{2}{c}{\cellcolor[HTML]{E0E0E0}}             \\
$\mathcal{M}_5 \ (y_{\mathrm{rot}})$   & \multicolumn{2}{c}{}                               & \multicolumn{2}{c}{\cellcolor[HTML]{E0E0E0}0.25} & \multicolumn{2}{c}{\cellcolor[HTML]{D0D0D0}0.44}     &                               & \multicolumn{2}{c}{\cellcolor[HTML]{E0E0E0}0.34}                    & \multicolumn{2}{c}{\cellcolor[HTML]{E0E0E0}3.77}     & \multicolumn{2}{c}{\cellcolor[HTML]{E0E0E0}(7.66)}       \\
$\mathcal{M}_6 \ (x_{\mathrm{rot}})$   & \multicolumn{2}{c}{}                               & \multicolumn{2}{c}{\cellcolor[HTML]{E0E0E0}0.16} & \multicolumn{2}{c}{\cellcolor[HTML]{D0D0D0}0.33}     &                               & \multicolumn{2}{c}{\cellcolor[HTML]{E0E0E0}0.17}                    & \multicolumn{2}{c}{\cellcolor[HTML]{E0E0E0}10.35}    & \multicolumn{2}{c}{\cellcolor[HTML]{E0E0E0}}             \\
                                       & \multicolumn{1}{l}{}        & \multicolumn{1}{l}{} & \multicolumn{1}{l}{}    & \multicolumn{1}{l}{}   & \multicolumn{1}{l}{}       & \multicolumn{1}{l}{}    &                               & \multicolumn{1}{l}{}                       &                        & \multicolumn{1}{l}{}                 &               & \multicolumn{1}{l}{}                  &                  \\
                                       & \multicolumn{6}{c}{\cellcolor[HTML]{FFFFFF}\textbf{Eigenvector of Linearized System $\bm{w}$}}                                                                                                     & \textbf{}                     & \multicolumn{6}{c}{\cellcolor[HTML]{D0D0D0}\textbf{Stepping in Hardware}}                                                                                                                                           \\
                                       & \multicolumn{1}{l}{\cellcolor[HTML]{FFFFFF}}        & \multicolumn{1}{l}{\cellcolor[HTML]{FFFFFF}} & \multicolumn{1}{l}{\cellcolor[HTML]{FFFFFF}}    & \multicolumn{1}{l}{\cellcolor[HTML]{FFFFFF}}   & \multicolumn{1}{l}{\cellcolor[HTML]{FFFFFF}}       & \multicolumn{1}{l}{\cellcolor[HTML]{FFFFFF}}    &                               & \multicolumn{2}{c}{\cellcolor[HTML]{D0D0D0}Forward Velocity [\si{\meter \per \second}]} & \multicolumn{4}{c}{\cellcolor[HTML]{D0D0D0}Frequency [\si{\hertz}]}                                                    \\
                                       & \multicolumn{1}{c}{\cellcolor[HTML]{FFFFFF}$x$}        & \multicolumn{1}{c}{\cellcolor[HTML]{FFFFFF}$y$} & \multicolumn{1}{c}{\cellcolor[HTML]{FFFFFF}$z$}    & \multicolumn{1}{c}{\cellcolor[HTML]{FFFFFF}roll}   & \multicolumn{1}{c}{\cellcolor[HTML]{FFFFFF}pitch}       & \multicolumn{1}{c}{\cellcolor[HTML]{FFFFFF}yaw}    &                               & \multicolumn{2}{c}{\cellcolor[HTML]{D0D0D0}}                        & \multicolumn{2}{c}{\cellcolor[HTML]{D0D0D0}dominant} & \multicolumn{2}{c}{\cellcolor[HTML]{D0D0D0}(underlying)} \\
$\mathcal{M}_1 \ (y_{\mathrm{trans}})$ & \multicolumn{1}{l}{\cellcolor[HTML]{FFFFFF}{[}0}    & \cellcolor[HTML]{FFFFFF}-0.27               & \cellcolor[HTML]{FFFFFF}0                       & \cellcolor[HTML]{FFFFFF}0.96                  & \cellcolor[HTML]{FFFFFF}0                          & \cellcolor[HTML]{FFFFFF}0$]^\tr$                &                               & \multicolumn{2}{c}{\cellcolor[HTML]{D0D0D0}0.038}                   & \multicolumn{2}{c}{\cellcolor[HTML]{D0D0D0}1.18}     & \multicolumn{2}{c}{\cellcolor[HTML]{D0D0D0}(3.63)}       \\
$\mathcal{M}_2 \ (x_{\mathrm{trans}})$ & \multicolumn{1}{l}{\cellcolor[HTML]{FFFFFF}{[}0.46} & \cellcolor[HTML]{FFFFFF}0                    & \cellcolor[HTML]{FFFFFF}0                       & \cellcolor[HTML]{FFFFFF}0                      & \cellcolor[HTML]{FFFFFF}0.89                       & \cellcolor[HTML]{FFFFFF}0$]^\tr$                &                               & \multicolumn{2}{c}{\cellcolor[HTML]{D0D0D0}-}                       & \multicolumn{2}{c}{\cellcolor[HTML]{D0D0D0}-}        & \multicolumn{2}{c}{\cellcolor[HTML]{D0D0D0}}             \\
$\mathcal{M}_3 \ (z_{\mathrm{rot}})$   & \multicolumn{1}{l}{\cellcolor[HTML]{FFFFFF}{[}0}    & \cellcolor[HTML]{FFFFFF}0                    & \cellcolor[HTML]{FFFFFF}0                       & \cellcolor[HTML]{FFFFFF}0                      & \cellcolor[HTML]{FFFFFF}0                          & \cellcolor[HTML]{FFFFFF}1$]^\tr$                &                               & \multicolumn{2}{c}{\cellcolor[HTML]{D0D0D0}0.084}                   & \multicolumn{2}{c}{\cellcolor[HTML]{D0D0D0}2.33}     & \multicolumn{2}{c}{\cellcolor[HTML]{D0D0D0}}             \\
$\mathcal{M}_4 \ (z_{\mathrm{trans}})$ & \multicolumn{1}{l}{\cellcolor[HTML]{FFFFFF}{[}0}    & \cellcolor[HTML]{FFFFFF}0                    & \cellcolor[HTML]{FFFFFF}-1                      & \cellcolor[HTML]{FFFFFF}0                      & \cellcolor[HTML]{FFFFFF}0                          & \cellcolor[HTML]{FFFFFF}0$]^\tr$                &                               & \multicolumn{2}{c}{\cellcolor[HTML]{D0D0D0}0.11}                    & \multicolumn{2}{c}{\cellcolor[HTML]{D0D0D0}3.60}     & \multicolumn{2}{c}{\cellcolor[HTML]{D0D0D0}}             \\
$\mathcal{M}_5 \ (y_{\mathrm{rot}})$   & \multicolumn{1}{l}{\cellcolor[HTML]{FFFFFF}{[}0}    & \cellcolor[HTML]{FFFFFF}0                    & \cellcolor[HTML]{FFFFFF}0                       & \cellcolor[HTML]{FFFFFF}0                      & \cellcolor[HTML]{FFFFFF}1                          & \cellcolor[HTML]{FFFFFF}0$]^\tr$                &                               & \multicolumn{2}{c}{\cellcolor[HTML]{D0D0D0}-}                       & \multicolumn{2}{c}{\cellcolor[HTML]{D0D0D0}-}        & \multicolumn{2}{c}{\cellcolor[HTML]{D0D0D0}}             \\
$\mathcal{M}_6 \ (x_{\mathrm{rot}})$   & \multicolumn{1}{l}{\cellcolor[HTML]{FFFFFF}{[}0}    & \cellcolor[HTML]{FFFFFF}0                    & \cellcolor[HTML]{FFFFFF}0                       & \cellcolor[HTML]{FFFFFF}1                      & \cellcolor[HTML]{FFFFFF}0                          & \cellcolor[HTML]{FFFFFF}0$]^\tr$                &                               & \multicolumn{2}{c}{\cellcolor[HTML]{D0D0D0}0.25}                    & \multicolumn{2}{c}{\cellcolor[HTML]{D0D0D0}6.80}     & \multicolumn{2}{c}{\cellcolor[HTML]{D0D0D0}}
\end{tabular}
\end{table*}
\subsection{Existence of modal oscillations on hardware}
\label{sec:results:modevalid}
To prove the practical relevance of the NNMs to develop locomotion in robotic hardware, predicted dynamics computed from the conservative \ebert model must be excitable and sustainable in the real system which is not energy-conservative and constantly loses energy, e.g., due to friction.
Thus, while NNMs are theoretically optimally efficient, sustaining them in hardware requires a controller to compensate for energy losses.
Therefore, we implemented a state-switching controller derived in previous works inspired by human control strategies \cite{lak14, manipulandum}.
The controller functions analogously to pushing a playground-swing when it naturally comes to a halt on either side, avoiding motor commands during the system's motion flow, which would disturb the natural dynamics.
Alternative approaches, such as feedback-modulated CPGs \cite{ijs23, ruppert2022nature} or even feedforward control given the known system frequencies could also be used to excite the oscillations. However, prior work comparing the alternatives \cite{sch21} showed the state-switching controller excites dynamics effectively with minimal tuning, motivating this choice for initial validation.

\begin{figure*}
    \centering
    \includegraphics[width=\textwidth]{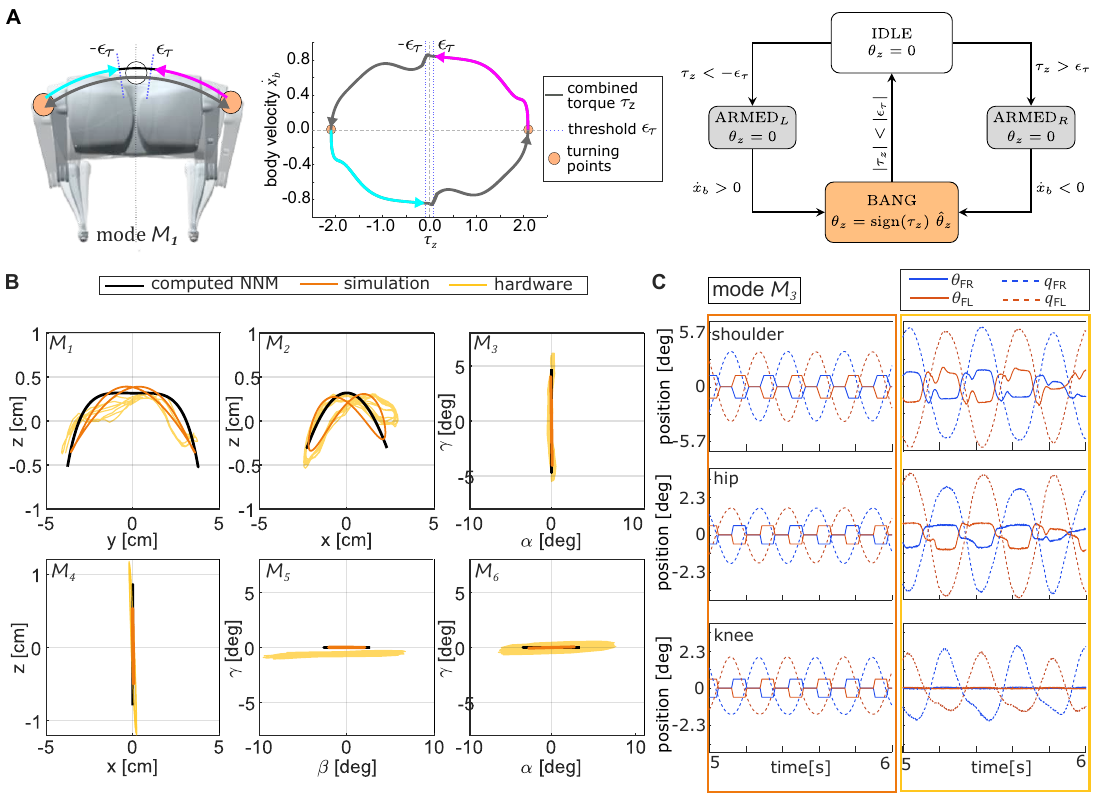}
    
    \caption{\textbf{Excitation of the NNMs in the hardware with a state switching controller.} \textbf{(A)} State machine of the switching control to sustain the modal oscillations in the simulation and hardware exhibiting friction, conceptually visualized for mode $\mathcal{M}_1$.
    \textbf{(B)} Overlay of the computed NNM of the conservative system with the Gazebo simulation and hardware system, which both incorporate friction effects. The motions of each mode $\mathcal{M}_1-\mathcal{M}_6$ are plotted on the respective relevant axes. \textbf{(C)} Example plot for the joints of one robot leg (top: shoulder, middle: hip, bottom: knee) taken from the oscillation of $\mathcal{M}_3$ in simulation (orange) and hardware (yellow). The commanded motor signals $\bm{\theta}$ and the robot's measured link position $\bm{q}$ are plotted for each joint, visualizing the amplification created by the resonance effects of the spring.
    }
    \label{fig:modes-comp}
\end{figure*}

The controller injects an empirically defined energy ${\theta}_z$ at the extreme points, where the robot's oscillation naturally subsides and the combined spring deflection $\tau_z$ \cite{lak14} is maximal. Crossing a threshold $\epsilon_\tau$ near equilibrium armed the control trigger (Fig.~\ref{fig:modes-comp}A), waiting to inject energy when the next direction change was detected from the combined Cartesian body velocity $\Dot{x}_b$.
The sign of $\theta_z$ depended on the oscillation direction according to
\begin{equation}
\label{eq:newbang-trigger}
\theta_z=
\begin{cases}
- \hat{\theta}_z  & \mathrm{if} \ \tau_{z} < -\epsilon_\tau \ \mathrm{and} \ \Dot{\bm{x}}_b < 0, \\
+ \hat{\theta}_z  & \mathrm{if} \ \tau_{z} > \phantom{-}\epsilon_\tau \ \mathrm{and} \  \Dot{\bm{x}}_b > 0, \\
0            & \mathrm{otherwise}. \\
\end{cases}
\end{equation}

The eigenvector $\bm{w}$ of each mode $\mathcal{M}_1-\mathcal{M}_6$ (Tab.~\ref{tab:linmodes-freq}, left) transformed the control signal back into joint space yielding a position command $\bm{\theta}_\mathrm{osc}$ for all joints (Eq.~\eqref{eq:theta_osc}).
This position jump was simultaneously commanded to all motors to momentarily inject energy (Fig.~\ref{fig:modes-comp}A). As the robot moves back towards the middle, $\tau_z$ decreases until reaching a minimum near the system equilibrium, where crossing $\epsilon_{\tau}$ armed or released the trigger depending on the motion direction (Fig.~\ref{fig:modes-comp}A).
Refer to the Methods section~\textit{\nameref{sec:method:bangbang}} for controller details, and Supplementary Material for chosen threshold values.

Applying the state-switching controller in a \ebert simulation with friction, and its hardware, successfully excited all six expected NNMs at distinct energy levels.
Modes $\mathcal{M}_3$–$\mathcal{M}_6$ remained  with their small oscillation amplitudes in a regime, where they appear linear, although Figure~\ref{fig:bert-modes} had verified that all but $\mathcal{M}_6$ become visibly nonlinear for higher energy levels. For the investigated regime, the simulations matched NNM predictions closely (Fig.~\ref{fig:modes-comp}B, orange), though hardware required higher energy due to more pronounced friction, resulting in slightly larger amplitudes (Fig.~\ref{fig:modes-comp}B, yellow).
Due to the lower oscillation frequencies, the modes $\mathcal{M}_1$ and $\mathcal{M}_2$ showed larger amplitudes entering the regime that exhibits pronounced nonlinear  behavior with hysteresis effects in both simulation and hardware.
This hysteresis may stem from a slight misalignment in the direction of energy injection at the turning points comparable to pushing the playground-swing off-center.
Nevertheless, the oscillations in \ebert remained stable, demonstrating the system's inherent self-stabilizing behavior. Minor asymmetries in $\mathcal{M}_2$ and offsets in $\mathcal{M}_5$ were most likely caused by hardware imperfections.

Comparing the mode frequencies of the ideal NNMs with simulation and hardware results (Tab.~\ref{tab:linmodes-freq}, left), showed that oscillation frequencies remained of similar magnitude across all cases, preserving the predicted sequence from $\mathcal{M}_1$ to $\mathcal{M}_6$.
However, oscillation frequencies in the simulation with friction were up to  $6 \%$ higher compared to the ideal conservative model, likely due to the control slightly altering the natural dynamics. In contrast, hardware frequencies were consistently lower (Tab.~\ref{tab:linmodes-freq}, left) likely because of unmodeled losses and actuation delays. The largest deviations (around 20\%) occurred for $\mathcal{M}_1$, $\mathcal{M}_3$, and $\mathcal{M}_6$, whose substantial lateral shoulder motion suggests increased joint friction or delay in this direction; all remaining modes stayed within 10\% of the predicted values.

Plotting the motor positions $\bm{\theta}$ versus link deflections $\bm{q}$ confirmed resonant amplification of power across all modes, exemplified  for mode $\mathcal{M}_1$ in Fig.~\ref{fig:modes-comp}C in simulation (left) and hardware (right).
Calculating the percentage $\eta$ of positive mechanical work provided passively per cycle further quantifies the exploitation of natural mechanical responses (Tab.~\ref{tab:linmodes-freq}, left) indicating a significant portion of the motion is passively generated, with some modes reaching $\eta$-values up to 80\%.

In summary, we confirmed that all six NNMs predicted for the conservative model of \ebert could be reliably excited and sustained in the hardware with minimal control largely preserving the system's intrinsic dynamics.

\subsection{From modes to locomotion}
\label{sec:results:locomodes}
After validating the predicted NNMs in the \ebert hardware, we finally test our hypothesis that each identified NNM may serve as foundation for a gait inspired by the biological concept of embodied intelligence \cite{van01, alexander2003principles}.
To this end, we gradually ramp up the energy injection with the state-switching controller to increase each mode amplitude until \ebert starts lifting its feet (Supplementary Video 1).
Commanding the free legs forward, we expected the robot to naturally \textit{fall} into gaits driven by resonances of its intrinsic dynamics.
Since NNMs are only defined for conservative, stable systems like the constrained standing pose analyzed previously, the NNM theory no longer strictly applies now.
Still, the modes capture dominant coordination patterns shaped by the robot's geometry, inertia, and elasticity - properties persisting despite contact changes. Thus, NNMs computed from the symmetric standing pose serve as proxies for the robot's natural movement tendencies.
This mirrors biological observations, where gaits reflect body resonance despite changing contacts \cite{van01, alexander2003principles}, motivating NNMs as foundation for locomotion beyond ideal conditions.
For the same reason, we relaxed the symmetric standing pose and considered more walking-oriented postures, where the COM is slightly shifted to aid falling forward \cite{joh22} and feet slightly turn outward to enhance stability \cite{Zink2020dogposture}.
Following a posture grid search (detailed in the Supplementary Materials), the \ebert's leg joint configuration for locomotion was set to
\begin{equation}
\begin{aligned}
    \bm{\theta}_{0,F} &= [-0.1, \ 0.4, -0.4]^\tr  \si{rad}, \quad \\
    \bm{\theta}_{0,H} &= [-0.1, \ 0.4, -0.3]^\tr  \si{rad} \ .
\end{aligned}
\label{eq:loco_config}
\end{equation}
The linearized eigenvectors were recomputed for the adapted leg configuration to guide the direction of the controlled energy injection and estimate the expected motion frequencies.
Consistent with earlier results (Tab.~\ref{tab:linmodes-freq}, left), the oscillation frequencies in hardware remained slightly lower than in simulation (Tab.~\ref{tab:linmodes-freq}, right).

\begin{figure*}[h!]
    \centering
    \includegraphics[width=\textwidth]{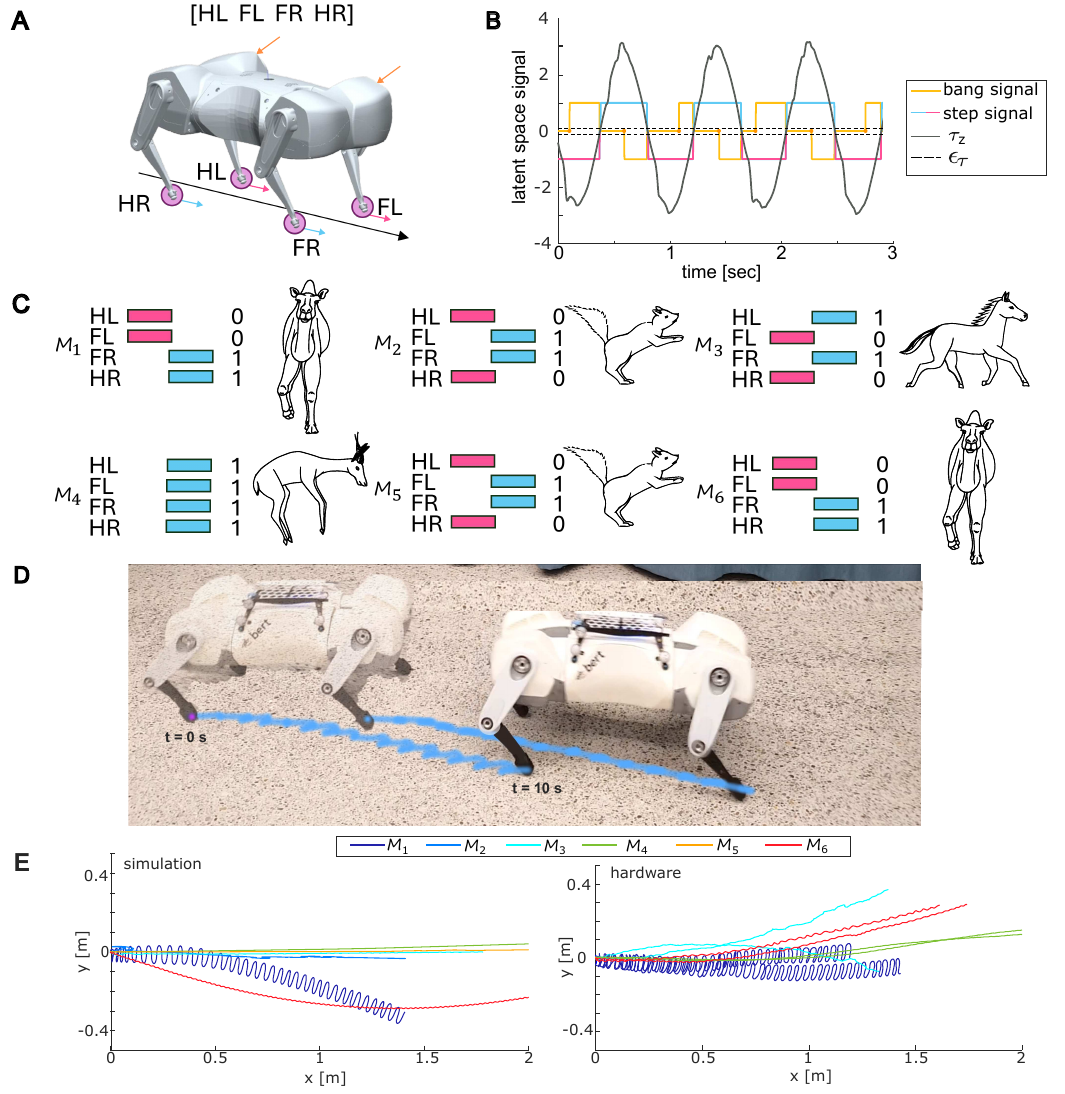}
    \caption{\textbf{Each NNM suggests a matching step pattern to develop a locomotion gait.} \textbf{(A)} Definition of the feet naming of the robot with arrows indicating the direction of the step trigger (blue/pink) and bang-signal (orange). \textbf{(B)} Control signals commanded in 1D-control space to trigger the bang-bang signal (yellow) and forward steps (blue/pink) based on the combined torque signal of all joints $\tau_z$ and the body velocity. \textbf{(C)}  For each mode, $\mathcal{M}_1-\mathcal{M}_6$, a matching step pattern can be identified, where a binary encoding can express which feet move together. \textbf{(D)} Visualization of the robot moving forward in a pace-like gait developed from $\mathcal{M}_1$ indicating the motion of the right foot pair. \textbf{(E)} Top view of the robot's COM moving forward is based on the oscillations of the different modes in simulation (left) and hardware (right).}
    \label{fig:loco-modes}
\end{figure*}

Gradually increasing energy during mode excitation in this new configuration, \ebert began to lift its feet in specific patterns in each mode (Fig.~\ref{fig:loco-modes}A, C). The foot pairs lifting simultaneously corresponded to symmetries in the linearized eigenvectors (Supplementary Tab.~\ref{tab:method:foot-pair}), suggesting that NNMs not only capture global body oscillations but also indicate which foot pattern aligns with each mode.
For example, the lateral rocking motions of $\mathcal{M}_1$ and $\mathcal{M}_6$ resembled the side-to-side swing of pace-gait seen in camels. The forward-backward motions of $\mathcal{M}_2$ and $\mathcal{M}_5$ aligned with simultaneous front-hind leg movements, typical of bounding in squirrels. Diagonal coordination in $\mathcal{M}_3$ suggested trotting as in horses, while vertical translation in $\mathcal{M}_4$ enabled hopping, observable in springboks (Fig.~\ref{fig:loco-modes}C).

To realize the forward stepping with the identified foot patterns, the state-switching controller was extended to command swing-leg motion whenever \ebert crossed the equilibrium, i.e., when weight shifted between foot pairs. The step command was retained until the next crossing, which triggered stepping of the opposite pair. Thus, the controller both sustained the mode oscillations and generated forward stepping (Fig.~\ref{fig:loco-modes}B, Supplementary Video). For controller details, see Methods \textit{\nameref{sec:method:steps}} and \textit{\nameref{sec:method:learn}}.

With the minimal control intervention, each intrinsic mode evolved into a gait in \ebert (Fig.~\ref{fig:loco-modes}C-E), where the respective gaits' forward velocity appeared to correlate with the frequency of the underlying mode (Tab.~\ref{tab:linmodes-freq}, right).
Interestingly, modes with differing frequencies but same foot pairing, like $\mathcal{M}_2$ and $\mathcal{M}_5$, produced different gait expressions: $\mathcal{M}_2$ resulted in slow bounding gait with alternating front-hind contact, while $\mathcal{M}_5$ produced a faster, hopping-like bound (see Supplementary Video 1).
An exception to the relation between mode frequency and forward velocity was only $\mathcal{M}_6$ in simulation, where the foot contact appeared too short to establish steps, causing the system to ``vibrate'' forward rather than walk.

In hardware, the modes $\mathcal{M}_1$, $\mathcal{M}_3$, $\mathcal{M}_4$, and $\mathcal{M}_6$ produced the predicted gaits from simulation, with $\mathcal{M}_6$ stabilizing into a gait likely due to friction increasing ground contact.
Overall, the gaits in hardware exhibited slightly reduced frequencies and step speeds compared to simulation (Tab.~\ref{tab:linmodes-freq}, right).
The modes $\mathcal{M}_2$ and $\mathcal{M}_5$, both involving motion about the y-axis, could not fully develop into gaits in hardware, because high torques in the hind legs' hip and knee joints caused unforeseen slippage in the belt transmissions. Although the actuators were designed to withstand these loads, the transmission slippage prevented sustained locomotion.
Notably, some gaits, like $\mathcal{M}_1$, exhibited two frequency peaks during stepping (Tab.~\ref{tab:linmodes-freq}, right).
In all gaits in simulation and the ones realizable in hardware, one peak aligned with the expected frequency of the pure modal oscillation, while the second frequency peak was likely induced by the step-driven energy input.

The combined findings support our hypothesis that intrinsic system dynamics, like NNMs, provide a promising foundation for generating distinct gaits in purposefully designed robots such as \ebert. Unique gaits naturally emerge from different modal oscillations, shaped by their frequency and direction.

\section{Discussion and Outlook}

Motivated by the biological principle of embodied intelligence, we explored whether nonlinear normal modes (NNMs) as a specific form of nonlinear dynamics may underlie naturally emerging gaits in robotic locomotion. To test this, we developed the elastic quadruped robot \ebert as a unique test platform encoding multiple slow, large-amplitude oscillations.
Established template-based approaches, such as the SLIP-inspired Raibert hoppers \cite{raibert1984hopping}, Thumper \cite{poulakakis2009spring}, ATRIAS \cite{hereid2014dynamic}, and Cassie \cite{reher2019dynamic}, exploit passive dynamics by shaping and stabilizing limit cycles within the Hybrid Zero Dynamics framework. In contrast, we pursued a fundamentally different perspective that engages the full mechanical system rather than relying on reduced-order templates.
This leads to dynamically consistent orbits that emerge naturally and are well-suited to stabilization via simple controllers, as we demonstrated in one-dimensional \cite{lak14} and multi-dimensional systems \cite{Calzolari2024tofix}.

The purposeful design of \ebert was enabled by our recently developed methods identifying six distinct NNMs within the 12-DOF robot, which could be excited in both simulation and hardware using a simple state-switching controller supporting the robot’s natural dynamics.
Remarkably, when this controller was extended to trigger steps in distinct foot pairs matching respective modal oscillations, on flat ground each NNM naturally unfolded into a gait without complex planning or precise foot placement in simulations, which largely translated to the hardware.

Motion amplification from motor to link (Fig.~\ref{fig:modes-comp}) and the metric $\eta$ (Tab.~\ref{tab:linmodes-freq}) quantified that mode resonances were successfully excited, indicating efficient use of elastic mechanics. While \ebert's stepping resembled biological gaits in temporal and spatial foot-contact symmetries, the gaits' energy efficiency was not yet evaluated nor were the relationship of arising stride length, forward velocity, and dynamic coupling, compared to biological or robotic counterparts.
Rather, this work showed that embodied intelligence in robots need not precisely replicate biological form, but intentionally shaping intrinsic dynamics can exploit this principle for functionally equivalent multi-gait emergence.
Nevertheless, future investigations on explicit control are needed to address how the different emerging gaits can be shaped to vary characteristics like step length and gait velocity along with performance evaluation of efficiency and stability across regimes and in more complex environments.

\subsection{\ebert as Test Platform for Embodied Intelligence}
\ebert presents a new unique platform implementing very high compliance encoding multiple nonlinear resonances in different directions, from which multiple diverse gaits developed expanding works on other bioinspired robots like BirdBot \cite{bad22}, Morti \cite{ruppert2022nature} and PAWS \cite{stella2025synergy}, where gaits were constrained to the sagittal plane.
Our robot design was guided by principles of Gan et al. \cite{Gan2014}, suggesting massless legs for energy conservative gaits.
Accordingly, \ebert's light legs (under 10\% of total mass) minimize impact losses during foot–ground contact and preserve the system’s intrinsic dynamics allowing the main body to dominate motion behaviors.
This mirrors biology, where fast-running animals like cheetahs have slender limbs supporting rapid repositioning.

However, robots lack biological tissue complexity and neuromuscular control to stabilize motions, making lightweight legs prone to vibrations during lift-off limiting passive dynamics exploitation, particularly complicated by soft SEA springs, during flight phases \cite{Calzolari2024tofix}.
Quasi-direct drives generating virtual elasticities could address this issue, allowing more precise foot coordination during flight since previous works confirmed that our dynamics analysis methods remain valid for stiff actuators, where NNMs and other periodic orbits similarly reduce energy expenditure \cite{bje22, Sachtler2024swingup, ehlert2026locomotion}. Nevertheless, \ebert intentionally embedded nonlinear dynamics through mechanical springs to minimize control and focus on the role of intrinsic mechanics in locomotion.

Notably, different from cheetahs and \ebert, many large terrestrial animals also feature heavier, proximally loaded limbs \cite{alexander1988elastic}, such as muscular horse thighs.
This anatomical configuration enhances stability and force production, by enabling greater inertial control during flight-stance transitions and reducing reliance on active control through momentum-driven passive stabilization \cite{stella2025synergy, biewener2006patterns}.
Thus, such mass distributions in robots could also enhance embodied intelligence, as demonstrated by BirdBot \cite{bad22} and PAWS \cite{stella2025synergy}, which incorporated biological leg mechanisms and weight distribution with actuation synergies leading to gaits that purposefully exploited leg dynamics. However, this approach demands detailed analysis of biology through methods like cadaver studies. Better general understanding biological principles and ways to embed them in robots without replicating specific biological mechanisms would be more scalable.
Our related work \cite{sesselmann2021embedding} demonstrated that NNMs can also be encoded in a system with heavier legs, which could facilitate smoother oscillation transitions and improved contact dynamics without relying on biological data. Future elastic quadrupeds could explore varied mass distributions to balance mechanical simplicity with bio-inspired effectiveness, which can be supported by our presented analysis tools.

\subsection{Applicability of Conservative NNMs to Robotic Hardware}
Our analysis methods could identify six NNMs in the conservative \ebert model with grounded legs, where especially the first two modes ($\mathcal{M}_1$, $\mathcal{M}_2$) showed pronounced nonlinearity with large deflections (Fig.~\ref{fig:bert-modes}) which is not predictable through linearization.
In theory, NNMs represent self-sustaining oscillations requiring no external energy; in practice, friction and motor dynamics distort system behavior, necessitating active control to drive oscillations. Evidently, excited modes in the simulation and hardware with friction shows discrepancies to the idealized conservative NNMs predicted with our tools (Fig.~\ref{fig:hardware}). Despite these deviations, prior work \cite{sch21,Calzolari2024tofix} and the \ebert experiments showed that the dynamics derived for the conservative system still dominate the behavior of the compliant quadruped even when applying minimal discrete control interventions.
Although improved methods will be needed for the transition from the ideal conservative model to real systems, NNMs prove their practical relevance and may serve as useful principle to shape and optimize intrinsic robotic dynamics \cite{sac22}.

\subsection{Limitations of Excitation Methods and Control Approach}
The original state-switching controller design \cite{lak14} injected energy near a system's equilibrium where linearized mode vectors characterize motion direction. However, applying this controller to the real \ebert hardware required adaptations as commanding the energy injection near the equilibrium caused slight disturbances in the motion flow due to the motors' delayed following of the instantaneous signals, which disturbed the natural dynamics. Tests showed that initiating the motor command at NNMs' turning points instead, where motions naturally come to a halt, reduced disturbance of the natural oscillations.

However, this adjustment caused a downside: linearized eigenvectors approximate energy injection direction only near equilibrium \cite{alb20, Sachtler2024swingup}, thus not remaining valid at the NNMs' turning points, especially for the highly nonlinear modes $\mathcal{M}_1$ and $\mathcal{M}_2$ at larger deflections. This caused observable hysteresis when exciting these modes in hardware (Fig.~\ref{fig:modes-comp}). Despite this, the simple state-switching controller remained effective due to the brief energy injection duration and the intrinsic mechanical dynamics dominating the motions.
Nevertheless, further ways to excite NNMs in hardware should be explored such as alternative control approaches like pure or feedback-modulated CPGs \cite{ijspeert2008central} or learning-based oscillators \cite{raffin2023simple}. Prior work had compared the differences of the state-switching controller versus different CPG-versions to excite modes in a single 2-DOF leg of an earlier \textit{bert}-robot \cite{sch21}, while ongoing research is currently comparing their performances on \ebert. Nevertheless, the state-switching controller remained a viable choice for the presented initial investigations because it allowed the robot's intrinsic mechanics to govern motions while primarily compensating friction.

\subsection{Potential of Gaits emerging from NNMs}
Regardless of control method for mode excitation (state-switching, CPGs, or learning-based approaches), NNMs seem to offer physics-grounded coordination priors linking structural compliance to feasible motion patterns.
In \ebert, foot-pair coordination that enabled forward stepping emerged directly from linearized eigenvector symmetries, with only step width learned via black-box optimization. The resulting gait shape appeared largely shaped by the underlying NNM, and the individual observations suggest that there may exist also a relationship between the gait velocity and frequency of the modal oscillations, consistent with biology where animals prefer different gaits for different speeds \cite{hoy81, alexander2003principles}. Therefore, the NNMs present a structural constraint naturally reducing coordination search spaces and aligning the robot’s mechanical DOFs, illustrating how modal structure can guide control strategies without prescribing them exclusively.

However, it is important to clarify that NNMs represent invariant manifolds that organize phase-space behavior around natural frequencies, rather than standalone motion generators. NNMs are formally only defined for smooth, conservative systems with fixed boundary conditions, while quadrupedal locomotion is inherently a hybrid process involving intermittent ground contact, friction, and state-dependent energy exchange. Consequently, a strict mathematical equivalence between the conservative NNM manifold and the observed hybrid gait attractors does not exist.
Instead, we conceptualize NNMs as intrinsic coordination templates encoding preferred phase relationships, symmetries, and joint-space trajectories.

Unlike reduced-order templates like the SLIP model, where gaits emerge explicitly from hybrid dynamics \cite{srinivasanRuina2006computer}, the conservative NNMs only provide a lower-dimensional morphological scaffold without replicating impact physics. Nevertheless, our experiments suggest that NNMs still serve as valid proxies for natural movement tendencies in the hybrid case. Related work on an alternative 8-DOF \textit{bert}-version demonstrated closer resemblance to the reduced-order templates drawing a formal link between conservative the modal structure and hybrid gaits by identifying and exciting periodic orbits across entire gait cycles \cite{Calzolari2024tofix}.

Consequently, our proposed framework covers neither all possible limit cycle types nor ways to actuate or control them;
rather, it shows how exploiting intrinsic dynamics in the form of NNMs, can reduce control burden and enhance robustness in compliant systems.

\subsection{Robotic Design aided by Intrinsic Dynamics}
Our research suggests that analyzing natural dynamics like NNMs during mechanical design offers new avenues for robot optimization by accounting for physical constraints and design limitations early on.
A key advantage of the NNM framework over linear modal analysis is its predictive power for large-amplitude, multi-directional dynamics. Linear modes only describe behavior near equilibrium, but tools like our NNM framework reveal how modes that may appear linear for small amplitudes like $\mathcal{M}3-\mathcal{M}6$ (Fig.~\ref{fig:modes-comp}) behave for larger energy-levels (Fig.~\ref{fig:bert-modes}). This aids designers to anticipate which modes will be stable and exploitable for intended motion patterns before building the robot.

Although experiments with \ebert verified that identified dynamics influence movement beyond conservative assumptions, they also revealed limitations of our applied design approach. By analyzing only the conservative model, we did not account for the full actuation chain, leading to belt slippage in $\mathcal{M}2$ and $\mathcal{M}5$ despite sufficient motor torque preventing us to realize these modes in hardware. This highlights the need for more robust methods to bridge from the analysis of idealized models to hardware. Thus, future implementations for automated co-design frameworks should treat mechanical dynamics as active learning priors \cite{ijspeert2013dynamical}. This can be supported by our  methods \cite{bje22, Sachtler2024swingup}, which were unavailable during \ebert’s early development stage, but now offer a powerful framework for such design analysis of robots when extended to include full actuation chains.

However, we emphasize whether dynamics like NNMs is \textit{functional} or \textit{efficient} remains fundamentally task-dependent.
While \ebert provides a compelling research platform for exploring embodied intelligence, its design highlights key trade-offs that may not suit commercial quadrupeds.
Although \ebert's high compliance enables diverse exploitable intrinsic dynamics with inherent impact resilience, the same compliance can reduce structural rigidity and complicate control.

While our work demonstrates the possibility to develop locomotion from intrinsic oscillations, it lacks rigorous evaluation across varying controllers, perturbations, and unstructured environments.
Future research is required for a systematic evaluation comparing elastic, dynamics-driven robots with conventional stiff-control systems, which needs quantifiable metrics to scope the potential and limitations of exploiting intrinsic dynamics. The analytical framework presented here can be extended for such comparisons as the methods are applicable to various architectures, including soft robots, tendon-driven systems, and gravity-driven rigid designs \cite{Sachtler2024swingup, pustina2024nonlinear}

In essence, the proposed methodologies suggest a promising direction for future robotics:
designing machines where mechanical structure and control co-evolve such that motion arises from physical intelligence rather than from independently scripted commands.
Future work aims to extend the NNM framework for task-driven co-design to parameterize the search space over leg compliance, mass distribution, and joint coupling by using NNMs or other intrinsic motion patterns as dynamic priors or stability constraints.
Learning from the limitations of our current platform, like the transmission constraints, nonlinear dynamics predicted by our methods should guide mechanical design decisions rather than only be applied to idealized models.
This can support automating discovery of morphology–controller pairs optimized for specific locomotion objectives (e.g., rough-terrain compliance, energy-minimized steady-state gaits, high-speed running). Aligning with embodied AI paradigms seeking to close the loop between physical design, intrinsic dynamics, and adaptive control, this could enable robots whose bodies and brains co-emerge through task-aware co-evolution \cite{gaba2026comprehensive}. Our work supports this vision by presenting a framework for robot design with predictable embodied motion patterns and demonstrating the applicability of NNM analysis to real hardware in the 12-DOF quadruped \ebert, where locomotion emerged from encoded mechanics with minimal control.

\section{Methods}
\label{sec:concept}

\subsection{Design for Purposeful Oscillations in \ebert}
\label{sec:method:stiffness}
\begin{figure*}
    \centering
    \includegraphics[width=0.88\textwidth]{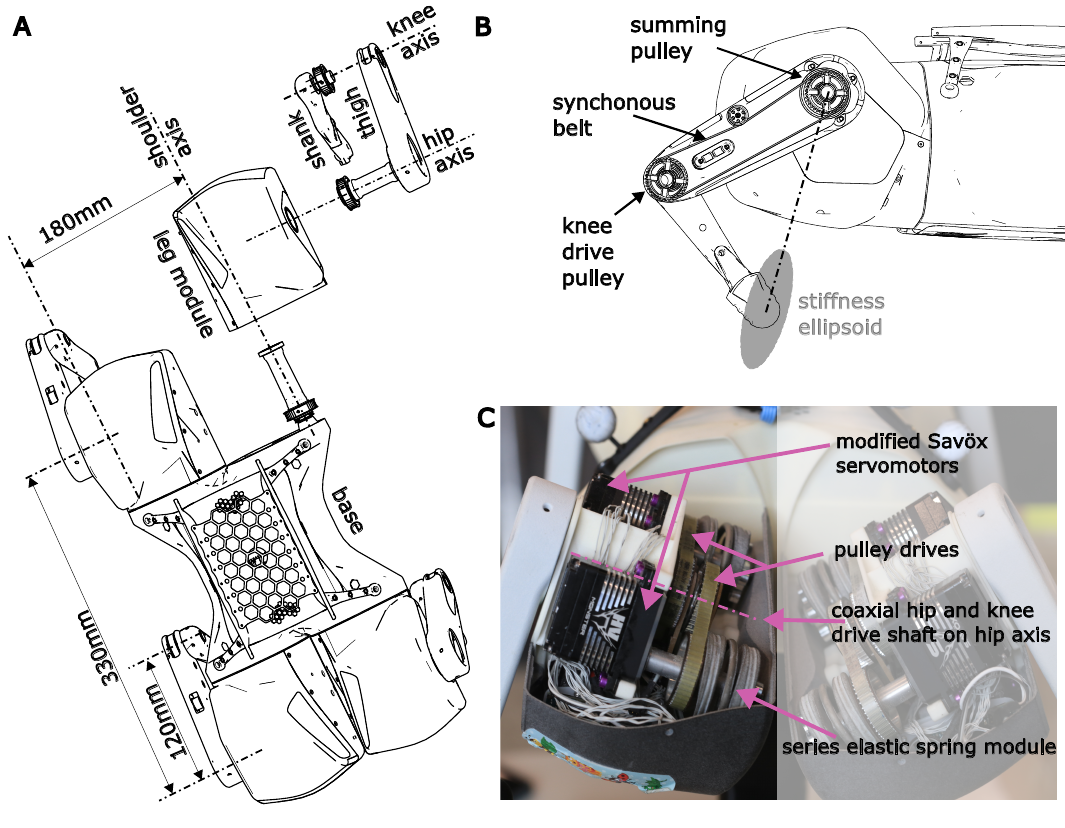}
    \caption{\textbf{Overview of the kinematics and drivetrain of the newly designed \ebert robot.} \textbf{(A)} Kinematic chain with rigid body base, leg module, thigh and shank, joint names, and key dimensions.
    \textbf{(B)} Orientation of the resulting stiffness ellipsoid with respect to the leg geometry as a result of the bi-articular coupling drivetrain transferring the summed motion of the hip joint and knee motor to the knee joint via a synchronous belt drive.
    \textbf{(C)} Drivetrain elements inside a leg module. }
    \label{fig:hardware}
\end{figure*}

The \ebert design followed previous versions of an 8-DOF \textit{bert}-quadruped to investigate intrinsic modal oscillations \cite{whitebert}, which were based on insights of \cite{Gan2014}.
Consequently, the robot kinematics and mass distribution were designed such that the foot tip elasticity in the sagittal plane corresponded to the assumptions in \cite{Gan2014}.
The robot dimensions were based on the body size of a small dog \cite{morey1992dogsize} with two leg segments, each having a length of \SI{120}{\milli \meter} (Fig.~\ref{fig:hardware}A).
The weight was estimated to be \SI{4.5}{\kilo \gram}, aiming to keep the total system mass low to minimize energy consumption. Based on these size and mass assumptions, the robot legs were designed. \\

\paragraph*{Kinematics}
\ebert was designed with a bi-articular coupled drivetrain for the hip and knee joints (Fig.~\ref{fig:hardware}B), tested in the previous \textit{bert} versions \cite{whitebert}.
This allowed the arrangement of the motors for the hip and knee off-axis in the shoulder module (Fig.~\ref{fig:hardware}A) to keep the mass of the legs minimal. This coupling also implemented an equal load sharing between the hip and knee motor. It aligned the principal axis of the stiffness ellipsoid with the legs' longitudinal axis (Fig.~\ref{fig:hardware}B).

\paragraph*{Stiffness Approximations}
\label{sec:method:robot}
Based on the size and mass estimations and the given kinematics, the stiffness of all joints of \ebert was estimated.
The compliance design is non-trivial since it determines the trajectories and frequencies of the natural oscillations.
Ideally, these oscillations should have frequencies with amplitudes that match the body dimension to facilitate effective and useful movement patterns.
During the design phase of \ebert, the methods to predict nonlinear oscillations from \cite{alb20, bje22, Sachtler2024swingup} were not yet fully developed and validated.
Therefore, we explored bio-inspired designs, where the leg stiffness of animals can be related to their speed, size, and body mass according to \cite[page~11, caption of Fig.~3]{farley1993jeb}.
\begin{equation}
    K_{\mathrm{leg}}=0.715M^{0.67\pm0.15}~\mathrm{kN \ m^{-1}} \ ,
\end{equation}
where $K_\mathrm{leg}$ is the combined leg stiffness of two legs assumed to be in contact, and $M$ refers to the mass of the system.

The estimated total mass $m=$~\SI{4.5}{\kilogram} led to a value of $K_\mathrm{leg} =$\SI{1.96}{\kilo \newton \per \meter}.
Projecting the stiffness in the knee with a leg segment length of $l=$~\SI{120}{mm} and assuming a normal knee angle $q_\mathrm{knee}=$~\SI{0.5}{\radian}, we obtained:
\begin{equation}
    K_\mathrm{knee}=K_\mathrm{leg} l^2 \sin^2(q_\mathrm{knee}) = \SI{6.48}{\newton\meter\per\radian} \;.
    \label{eq:K_Knee}
\end{equation}
Respecting the actuation kinematics, the closest available spring stiffness of $K = K_\mathrm{knee} = K_\mathrm{hip} = $ \SI{6.55}{\newton\meter\per\radian} was implemented in the robot's hips and knees.

To estimate the shoulder stiffness, we analyzed the reflected Cartesian stiffness of the overall system and assessed its local stability. The analysis assumed a configuration where the shoulders remained at the neutral position, ensuring no static torque on them, with the leg springs $K_\mathrm{hip}, K_\mathrm{knee}$ and their corresponding rest positions $q_\mathrm{hip}, q_\mathrm{knee}$.
Under the effect of gravity, this setup led to an equilibrium pose $\x_\mathrm{eq}$ of the trunk at a height of \SI{18.2}{cm}, with hip and knee joint angles of approximately \SI{0.77}{rad}.
With $\q$ denoting the joint angles, we expressed the reflected Cartesian stiffness $\K_{x} \in \mathbb{R}^{6\times6}$ \cite{Chen2000} in trunk coordinates as
\begin{equation}
    \small
    \K_x = \Jac_{qx}^\tr \K_q \Jac_{qx} \;, \quad \K_q = \left.\frac{\partial^2 V}{\partial \q^2}\right|_{\q_\mathrm{eq}} \;, \quad \Jac_{qx} = \left.\frac{\partial f_\mathrm{ik}}{\partial \x}\right|_{\x_\mathrm{eq}} \ ,
\end{equation}
where $\q_\mathrm{eq} = f_\mathrm{ik}(\x_\mathrm{eq})$ represents the inverse kinematics solution with fixed feet, and $\Jac_{qx}$ its Jacobian.
A linear stability analysis indicated that achieving a marginally stable equilibrium at $\x_\mathrm{eq}$, characterized by eigenvalues with real parts approximately zero, necessitates a shoulder stiffness of at least \SI{4}{\newton \meter \per \radian}. To achieve a similar reflected Cartesian stiffness in the X and Y directions, i.e., $K_{x}^{(1,1)} \approx K_{x}^{(2,2)}$, a shoulder stiffness of approximately \SI{9.6}{\newton\meter\per\radian} is required. Increasing the shoulder stiffness beyond this value raises the stiffness in the Y direction, which promotes sagittal motion rather than excessive lateral oscillations.
Thus, to obtain symmetric rotational stiffness in the shoulders, guarantee stability, and encourage forward locomotion, we chose \SI{13}{\newton\meter\per\radian} for the shoulder springs resulting in a stiffness matrix for all joints of
\begin{equation}
    \K_{\q}=\begin{bmatrix}13.1 & -6.55 \\ -6.55 & 6.55 \end{bmatrix} \ .
\end{equation}

\paragraph*{Energy considerations}
To estimate the robustness of the \ebert system, the mechanical energy storage capability of the springs was considered.
The equal distribution of gravitational forces on all joints in the sagittal plane, i.e., hip and knee joints, allows the assumption that all the engaged springs contribute equally to vertical energy storage. This entails that all the energy stored in the springs of the knee and hip joints, here denoted $K=$~\SI{6.55}{\newton\meter\per\radian}, can be released to counteract gravity:
\begin{equation}
    mgh = \frac{1}{2K} n \tau_{\mathrm{max}}^2
\end{equation}
where $\tau_{\mathrm{max}}$ is the joint max torque in Tab.~\ref{tab:spec}.
Considering all joints acting in the sagittal plane ($n=8$) gives an estimate of the robustness of the system against falling:
\begin{equation}
    h= \frac{\tau_{\mathrm{max}}^2 n}{K m g} \ .
\end{equation}
Substituting the relevant values for mass and stiffness revealed that \ebert should be robust to impacts from \SI{59.6}{\centi \meter} without exceeding the allowable torque.

Details on the realization of the system design in robotic hardware can be found in the Supplementary Methods \textit{\nameref{sec:app:robot}}. The total weight of the completed robot of \SI{4.56}{\kilo\gram} matched well the initial bioinspired approximations. However, when adding the battery of \SI{0.65}{\kilo\gram}, the completed system was slightly heavier, with \SI{5.2}{\kilo\gram}. All final dimensions of \ebert are summarized in Fig.~\ref{fig:hardware}A and the Supplementary Table~\ref{tab:spec}.

\subsection{Simulation model and pipeline}
For initial validation and testing, a multibody system model was derived from the CAD data of the \ebert hardware to simulate the system using Gazebo 11 with the default ODE solver.
For the initial mode excitation, the robot's feet were constrained to the ground by ball joints, while later these joints were removed such that the robot could lift its feet to develop locomotion.
No dedicated motor model was implemented in the robot simulation. Still, a velocity limiter was applied to the motor signal in simulation and hardware alike, such that the behavior closely matched.
The model inputs and outputs were defined according to the hardware's available interfaces so that an identical control framework implemented in \textit{MATLAB Simulink} could be used.
The motor position control of the SEA-drive train in hardware was simulated with a PD-controller, where the P-gain corresponded to the spring stiffness $k_j$ in each joint $j$, and the D-value applied a damping $d_j = 0.02k_j$ to approximate hardware friction effects. Thus, commanded motor positions were relayed to the simulated robot joints analogous to the SEA-driven hardware.
A deflection limiter was also applied in hardware and simulation to protect motors from physical overloads above the \SI{5}{\newton \meter} set as a limit during the design phase.
Positions and velocities of the motors and links were available as feedback in simulation and hardware.
As middleware, the DLR-developed \textit{Links and Nodes} was used \cite{schmidt2026links}.

\subsection{Nonlinear modal analysis}
\label{sec:method:nnm}
To identify the nonlinear modes of the quadruped robot, we utilize a conservative model based solely on the interaction of multi-body dynamics with gravitational and elastic forces.
To find the modes of the mechanical system, we remove the motor dynamics by fixing the motor positions to the rest configuration of the springs, and we fix the feet to the ground. Starting from the eigenvectors of the linearized system, we find the nonlinear modes of the quadruped using numerical continuation.
The procedure is detailed in the section \textit{\nameref{sec:app:nnm}} of the Supplementary Material.

\subsection{Mode excitation controller}
\label{sec:method:bangbang}
NNMs extend the concept of linear normal modes, which characterize motions in linear systems or small-amplitude oscillations in nonlinear ones. In linear modes, all components oscillate at the same natural frequency and with a constant phase relationship.
Exciting modes at their natural frequency induces resonance, which efficiently transfers energy into the system, leading to an increasing amplitude.
Local linear modes of \ebert\ can be identified by linearizing the equations of motion around an upright standing position, yielding a multi-dimensional second-order system decomposed into $n$ modal oscillators. From these, our methods can compute their nonlinear extensions by numerically continuing solutions as energy increases. This extends the concept of modes as a ``vibration-in-unison'' of all masses \cite{ROSENBERG1966}. Details for computing the NNMs of \ebert can be found in the Supplementary Material section \textit{\nameref{sec:app:nnm}}.

The controller to excite and sustain the predicted NNM oscillations in a hardware system with friction was based on our previous research that deduced a simple switching control law from observations of human strategies to induce intrinsic system oscillations \cite{lak14}. The control approach allowed for the orchestration of multiple joints with a single signal by linearly transforming sensory input from a multi-dimensional joint space into a one-dimensional controller space \cite{lak14}. The needed transformation weights $\bm{w}$ could be extracted with a learning rule that was mathematically equivalent to \textit{Oja's rule}, which showed that the weights converged to the mode vector of the linearized system for close-to-optimal energy-efficient movements \cite{stra16}. It was validated in various elastic robots, a biomimetic robotic arm \cite{lak14}, and a hopping leg \cite{stra16, wandinger2024tele}, as well as a previous \textit{bert} version \cite{seidel2024toward}.
Additionally, simulations \cite{stra16b} and a human subject study supported the hypothesis of a similarly functioning effect in humans to control dynamic body movements \cite{manipulandum}.

Applying this control law to the 12-DOF \ebert, required to calculate the torques $\bm{\tau}$ in each joint based on the measured motor positions $\bm{\theta}$ and link positions $\bm{q}$ using the known spring stiffness matrix $\bm{K}$:
\begin{equation}
    \bm{\tau} = \bm{K_q}(\bm{\theta}-\bm{q}) \ .
\end{equation}
Using the eigenvectors $\bm{w}$ of the linearized system for the different modes $\mathcal{M}_{1-6}$ as stated in Tab.~\ref{tab:linmodes-freq}, the individual joint torques can be mapped into a one-dimensional control space expressed as the scalar value $\tau_z$ that combines all joint torques:
\begin{equation}
\label{eq:tau_z}
    \tau_{z} = \frac{1}{||\bm{w}||} \bm{w}^\mathsf{T} \bm{\tau} \ .
\end{equation}
Within this control space, the trigger signal $\hat{\theta}_z$ was set according to Equation~\eqref{eq:newbang-trigger}, which detected the point on which the state-switching controller should inject energy by monitoring the combined Cartesian body velocity $\Dot{x}_b$, which flipped the sign at the return points where the oscillation in one direction naturally subsided. The vector $w$ corresponded to the respective mode vector of the linearized system listed in Table~\ref{tab:linmodes-freq} and the chosen parameter values for $\hat{\theta}_z$ and $\epsilon_\tau$ can be found in the Supplementary method section \textit{\nameref{sec:app:parameters}}.

Applying the vectors $\bm{w}$ of the linearized modes as transformation weights again, $\hat{\theta}_z$ was mapped back to joint space, providing a position command $\bm{\theta}_{\mathrm{osc}}$ for all joints:
\begin{equation}
\label{eq:theta_osc}
    \bm{\theta}_{\mathrm{osc}} = \frac{\bm{w}}{||\bm{w}||} \theta_z \ .
\end{equation}
This position command was added to the initial position of the motors $\bm{\theta}_0$ that combines the joint commands to the individual legs $\bm{\theta}_0 = [\bm{\theta}_{\mathrm{FR},0}, \bm{\theta}_{\mathrm{FL},0}, \bm{\theta}_{\mathrm{HR},0}, \bm{\theta}_{\mathrm{HL},0}]$. Thus, the combined signal $\bm{\theta}_{\mathrm{cmd}}$ to sustain the mode oscillations became
\begin{equation}
    \bm{\theta}_{\mathrm{cmd}} = \bm{\theta}_0 + \bm{\theta}_{\mathrm{osc}} \ .
\end{equation}

\subsection{Quantifying the exploitation of nonlinear resonance}
To quantify the contribution of the natural mechanical response to achieve motion, we define the efficiency $\eta$ as the ratio of positive work performed by the actuator to the positive joint work \cite{Hutter2013efficient}
\begin{equation}
\label{spring_eta}
    \eta = 1 - \frac{\int^T \max(\dthm^\tr\bs{\tau},0) \ \mathrm{d}t}{\int^T \max(\dq^\tr\bs{\tau},0) \ \mathrm{d}t} \;.
\end{equation}
This indicates the percentage of positive work that can be provided passively in a cycle.
A value of 1 corresponds to perfect exploitation of the nonlinear passive dynamics, while a value of 0 means that the control is not taking advantage of the mechanics.
It was used in \cite{Hutter2013efficient} to assess mechanical efficiency in single-legged robotic hopping, and it can also serve to compare performance across different robots and against animal in-place hopping.
Applied to \ebert, the calculated values suggest that the nonlinear passive dynamics were exploited across all modes in simulation and hardware, although with varying percentage (Tab.~\ref{tab:linmodes-freq}).

\subsection{Initial joint configurations}
To initially analyze and validate the NNMs in \ebert, a symmetric configuration was chosen, commanding identical motor positions to all joints per leg in \ebert as stated in Equation~\eqref{eq:leg_angle_def}.
However, as stated in the Results section \textit{\nameref{sec:results:locomodes}}, a more natural pose was assumed to discover locomotion based on the known characteristics of dogs, where the COM is slightly shifted to the front and the feet are turned outwards \cite{joh22}. For such poses, a grid search over varying initial configurations was carried out, always assuming that all feet remained attached to the ground. As expected, the dedicated NNM shape and frequencies varied with the different configurations; however, the overall motion patterns and frequencies of the identified modes $\mathcal{M}_{1-6}$ always remained similar. Applying the aforementioned state-switching controller to the different configurations in simulations revealed that for some configurations, the inclined robot body pose was already sufficient to realize a forward motion without adding steps. Taking the covered distance in simulation for each mode as a metric to judge the suitability of a configuration for locomotion, eventually, the joint configuration stated in Equation~\eqref{eq:loco_config} was determined as most promising. Details on the grid search can be found in the Supplementary Methods \textit{\nameref{sec:app:grid}}.

\subsection{Encoding of modal feet couplings}
\label{sec:method:couple}
For each modal oscillation, a foot pairing that naturally matched the intrinsic motion could be extracted from symmetries of the linearized eigenvectors in joint space (Supplementary Tab.~\ref{tab:method:foot-pair}). The derived pairing of feet that should move simultaneously based on these symmetries per mode (Fig.~\ref{fig:loco-modes}B) could be expressed through a binary vector. Thus, the feet that were derived to move together were encoded with a $1$, while the other foot pair was encoded with $0$. For example, the motion along the $y$-axis manifested in the modes $\mathcal{M}_1$ and $\mathcal{M}_6$ (Fig.~\ref{fig:bert-modes}C) suggested moving the two feet on the left simultaneously in an alternate fashion with steps from the two right feet. The corresponding binary stepping vector $b_\mathrm{step}$ was thus:
\begin{equation}
    [\mathrm{HL, FL, FR, HR}] = [0, \ 0, \ 1, \ 1] = b_\mathrm{step} (m_1)\ .
\end{equation}
The motions of $\mathcal{M}_2$ and $\mathcal{M}_5$ around the y-axis proposed to move the front feet pair alternating with the back feet, leading to a bounding-like gait
($b_\mathrm{step} = [0, \ 1, \ 1, \ 0] $).  %
The rotation around the $z$-axis realized by mode $\mathcal{M}_3$ suggested the diagonal feet pair to be simultaneously weight-bearing, similar to the step pattern in trotting horses. Thus, the front right leg is moved with the hind left leg, and the front left with the hind right leg
($b_\mathrm{step} = [1, \ 0, \ 1, \ 0] $). %
Exciting the fourth mode $\mathcal{M}_4$ naturally led to a hopping motion, where all feet lifted from the ground simultaneously. Here, the robot's body inclination sufficed to create a forward motion, such that no dedicated step had to be commanded for this mode
($b_\mathrm{step} = [1, \ 1, \ 1, \ 1] $).

\subsection{Stepping Controller}
\label{sec:method:steps}
The state-switching controller was extended to trigger the alternating forward stepping of the identified foot pairs for each modal oscillation of \ebert to develop locomotion patterns.
As in Equation~\eqref{eq:newbang-trigger}, this step command $s_{feet}$ was triggered in the 1D control space to apply a step whenever the robot shifted its weight from one foot-pair to the other. As trigger value, the earlier defined $\tau_z$ metric was used that combined the torques of all joints according to Equation~\eqref{eq:tau_z}. Since the combined torques should be at a minimum close to the equilibrium position, the same threshold value $\epsilon_{\tau}$ of the initial state switching controller could be used to detect the zero-crossing where the robot's weight support shifted from one foot-pair to the other.
However, in contrast to the switching rule to inject energy in Equation~\eqref{eq:newbang-trigger}, the step signal was applied independently of the Cartesian body velocity.
Instead, crossing $\epsilon_{\tau}$ directly triggered the step command $s_{\mathrm{feet}}$ that encoded which leg pair should move forward. This step command had to be sustained until the robot next crossed the equilibrium position, where the robot's weight shifted back to the feet that had previously stepped forward, freeing the opposite foot-pair.
Thus, the corresponding switching rule was defined as
\begin{equation}
\label{eq:steptrigger}
s_{\mathrm{feet}} =
\begin{cases}
b_\mathrm{step} & \mathrm{if} \ \tau_{z} > \epsilon_\tau \ \\
!b_\mathrm{step} & \mathrm{if} \ \tau_{z} < -\epsilon_\tau \ \\
[0, \ 0, \ 0, \ 0 ]            & \mathrm{otherwise} \\
\end{cases} \ ,
\end{equation}
The step vector $s_{\mathrm{feet}}$ was multiplied with the determined step length encoded by defined values for the shoulder, hip, and knee joint of each individual leg, leading to a $4\times 3$ matrix. Each row encoded the position change for the three joint motors of each leg to command a step, where the order of the rows remained $[\mathrm{FR, FL, HR, HL}]^\tr$. Through the binary encoding, only one foot-pair was commanded to move, while the zeros cancel the step term for the stance legs.
Flattening the obtained matrix, the step positions $\bm{\theta}_\mathrm{step}$ were added to the previously derived motor control for a combined position command of
\begin{equation}
    \bm{\theta} = \bm{\theta}_0 + \bm{\theta}_{\mathrm{osc}} + \bm{\theta}_{\mathrm{step}} \ .
\end{equation}
The applied parameter values for $\hat{\theta}_z$ and $\epsilon_\tau$ with each mode can be found in the Supplementary method section \textit{\nameref{sec:app:parameters}}.

\subsection{Optimizing step length}
\label{sec:method:learn}
To determine the step length fitting each modal oscillation, we used black-box optimization.
The shoulder joint was assumed to remain in its plane, as stepping forward does not require leg abduction. Only the values for the hip and knee joints were changed to achieve a forward step.
To further limit the search space, we took advantage of the robot's sagittal symmetry, assuming the left and right legs would take the same step.
However, since the robot was inclined forward, we optimized the front and hind legs separately. This resulted in optimizing two parameters for the front legs ($F$) and two for the hind legs ($H$), giving a search space of four parameters:
\begin{equation}
    \bm{\theta}_{F,\mathrm{step}} = [0, h_{F}, k_{F}]^\tr; \ \bm{\theta}_{H,\mathrm{step}} = [0, h_{H}, k_{H}]^\tr \ ,
\end{equation}
where the parameters $h$ and $k$ refer to the hip and knee joints, respectively.
In addition to optimizing the step length, we also relearned the scaling of the motor signal $\hat{\theta}_z$ since the added energy and motion from stepping forward required a smaller energy injection at the turning points compared to the value to maintain pure oscillations.

Empirically estimated values for the parameters were assumed to start the optimization process. These values were optimized in simulation, where the robot was repeatedly initialized and excited in the different mode oscillations.
Next, the extended state switching controller was activated to add a step, starting the learning trial.
Each simulation trial lasted \SI{8}{\second}, with the forward velocity serving as the objective function~\cite{raffin2023simple}.
The used algorithm for the optimization was the covariance matrix adaptation evolution strategy (CMAES) \cite{hansen2001completely} from the Optuna library \cite{takuya2019optuna}.

For all modes, it took 100-200 simulation trials to find suitable step parameters that led to gait patterns realizing locomotion forward. Following, the optimization process for each mode was repeated directly on the hardware, starting from the simulation values. The determined step values for each mode can be found in the Supplementary method section \textit{\nameref{sec:app:parameters}}.

\section{Data and materials availability}
All data needed to evaluate the conclusions of the paper are available in the paper or the Supplementary Materials. Additionally, the raw data of the robot trials in simulation and hardware are deposited on figshare accessible under \\
\href{https://doi.org/10.6084/m9.figshare.29514518}{https://doi.org/10.6084/m9.figshare.29514518} \cite{ebert-figshare}.

\section{Code availability}
The code to analyze the provided data stated above can likewise be found in the same figshare repository accessible over
\href{https://doi.org/10.6084/m9.figshare.29514518}{https://doi.org/10.6084/m9.figshare.29514518}.
In this repository, you find the raw data recorded for the robot experiments in simulation and hardware as well as the nonlinear normal modes computed for the \ebert robot using the methods detailed in the Supplementary Materials. Additionally, we provide code to recreate the figures of the manuscript.
Furthermore, we provide separately the code to calculate the oscillation modes of the \ebert robot as detailed in the Supplementary Materials under \href{https://github.com/anschm189/eBert_modes.git}{https://github.com/anschm189/eBert\_modes.git} \cite{ebert-git}. This repository visualizes the nonlinear normal modes of the quadruped model of \ebert.
While the underlying continuation algorithm was developed by Yannik P. Wotte, Filip Bjelonic, Cosimo Della Santina, and Arne Sachtler in previous work, the contribution of this manuscript is the \ebert-compatible quadruped model and the code for its analysis and simulation developed by Davide Calzolari, which are new and specific to this work. The repository contains the full model and analysis pipeline for eBert.

\section{Acknowledgements} The authors thank Yannik P. Wotte (University of Twente, Bergische Universität Wuppertal), Filip Bjelonic (ETH Zurich), and Dr. Cosimo Della Santina (TU Delft) for their support in developing algorithms used for the nonlinear modal analysis.
Furthermore, the authors thank Dr. Michael Panzirsch and Dr.-Ing. Alexander Dietrich (both DLR) for internal feedback on early manuscript versions. They also thank Nepomuk Werner for supporting the final code preparation. 

\section{Funding statement} This research was supported by the European Research Council (ERC) through the European Union’s Horizon 2020 Research and Innovation Programme under Grant 835284 (M-Runners).

\section{Author contributions statement}
\noindent
\begin{tabularx}{\columnwidth}{lX}
	Conceptualization:& {\small F.L., D.S., M.H., T.G., A.A-S., R.B., D.C., An.S.; }\\
	Methodology:& {\small Ar.S., A.R., F.L., D.S., M.H., T.G., A.A-S., D.C., An.S., M.P.;} \\
	Investigation:& {\small F.L., D.S., M.H., T.G., D.C., R.B., F.S., T.E., A.R., An.S.;} \\
	Visualization:&  {\small A.R., An.S., Ar.S., M.P.;}\\
	Funding acquisition:& {\small A.A-S., D.S., F.L.;}\\
	Project administration:& {\small An.S., A.A-S.;} \\
	Supervision:& {\small A.A-S.;} \\
	Original draft:& {\small An.S., D.C., F.L.;} \\
	Review and editing:& {\small An.S., Ar.S., D.C., F.L., M.K., D.W., M.P., D.S., A.R., T.E., J.L., A.A-S.}
\end{tabularx}

\section{Competing interests} The authors declare that they have no competing interests.

\bibliographystyle{IEEEtran}

\begin{thebibliography}{10}
\providecommand{\url}[1]{#1}
\csname url@samestyle\endcsname
\providecommand{\newblock}{\relax}
\providecommand{\bibinfo}[2]{#2}
\providecommand{\BIBentrySTDinterwordspacing}{\spaceskip=0pt\relax}
\providecommand{\BIBentryALTinterwordstretchfactor}{4}
\providecommand{\BIBentryALTinterwordspacing}{\spaceskip=\fontdimen2\font plus
\BIBentryALTinterwordstretchfactor\fontdimen3\font minus
  \fontdimen4\font\relax}
\providecommand{\BIBforeignlanguage}[2]{{%
\expandafter\ifx\csname l@#1\endcsname\relax
\typeout{** WARNING: IEEEtran.bst: No hyphenation pattern has been}%
\typeout{** loaded for the language `#1'. Using the pattern for}%
\typeout{** the default language instead.}%
\else
\language=\csname l@#1\endcsname
\fi
#2}}
\providecommand{\BIBdecl}{\relax}
\BIBdecl

\bibitem{bis21}
P.~Biswal and P.~K. Mohanty, ``Development of quadruped walking robots: A
  review,'' \emph{Ain Shams Engineering Journal}, vol.~12, no.~2, pp.
  2017--2031, 2021.

\bibitem{chi23}
O.~H. Chi, C.~G. Chi, D.~Gursoy, and R.~Nunkoo, ``Customers’ acceptance of
  artificially intelligent service robots: The influence of trust and
  culture,'' \emph{International Journal of Information Management}, vol.~70,
  p. 102623, 2023.

\bibitem{fuk22}
A.~Fukuhara, M.~Gunji, and Y.~Masuda, ``Comparative anatomy of quadruped robots
  and animals: a review,'' \emph{Advanced Robotics}, vol.~36, no.~13, pp.
  612--630, 2022.

\bibitem{zef03}
A.~Zeffer, L.~C. Johansson, and {\AA}.~Marmebro, ``Functional correlation
  between habitat use and leg morphology in birds (aves),'' \emph{Biological
  Journal of the Linnean Society}, vol.~79, no.~3, pp. 461--484, 2003.

\bibitem{niven2008}
J.~E. Niven and S.~B. Laughlin, ``Energy limitation as a selective pressure on
  the evolution of sensory systems,'' \emph{Journal of Experimental Biology},
  vol. 211, no.~11, pp. 1792--1804, 2008.

\bibitem{valero2022bio}
F.~J. Valero-Cuevas and A.~Erwin, ``Bio-robots step towards brain--body
  co-adaptation,'' \emph{Nature Machine Intelligence}, vol.~4, no.~9, pp.
  737--738, 2022.

\bibitem{pfeifer2007self}
R.~Pfeifer, M.~Lungarella, and F.~Iida, ``Self-organization, embodiment, and
  biologically inspired robotics,'' \emph{science}, vol. 318, no. 5853, pp.
  1088--1093, 2007.

\bibitem{Brooks1991aiBody}
R.~A. Brooks, ``New approaches to robotics,'' \emph{Science}, vol. 253, no.
  5025, pp. 1227--1232, 1991.

\bibitem{blickhan2021trunk}
R.~Blickhan, E.~Andrada, E.~Hirasaki, and N.~Ogihara, ``Trunk and leg
  kinematics of grounded and aerial running in bipedal macaques,''
  \emph{Journal of Experimental Biology}, vol. 224, no.~2, p. jeb225532, 2021.

\bibitem{van01}
J.~P. van~der Weele and E.~J. Banning, ``Mode interaction in horses, tea, and
  other nonlinear oscillators: The universal role of symmetry,'' \emph{American
  journal of physics}, vol.~69, no.~9, pp. 953--965, 2001.

\bibitem{kur08}
Y.~Kurita, Y.~Matsumura, S.~Kanda, and H.~Kinugasa, ``Gait patterns of
  quadrupeds and natural vibration modes,'' \emph{Journal of System Design and
  Dynamics}, vol.~2, no.~6, pp. 1316--1326, 2008.

\bibitem{hoy81}
D.~F. Hoyt and C.~R. Taylor, ``Gait and the energetics of locomotion in
  horses,'' \emph{Nature}, vol. 292, no. 5820, pp. 239--240, 1981.

\bibitem{alexander2003principles}
R.~M. Alexander, \emph{Principles of animal locomotion}.\hskip 1em plus 0.5em
  minus 0.4em\relax Princeton university press, 2003.

\bibitem{Cavagna2015running}
G.~A. Cavagna and M.~A. Legramandi, ``Running, hopping and trotting: tuning
  step frequency to the resonant frequency of the bouncing system favors larger
  animals,'' \emph{Journal of Experimental Biology}, vol. 218, no.~20, pp.
  3276--3283, 2015.

\bibitem{alexander1988elastic}
R.~M. Alexander, \emph{Elastic mechanisms in animal movement}.\hskip 1em plus
  0.5em minus 0.4em\relax Cambridge University Press, 1988.

\bibitem{del20}
C.~Della~Santina, M.~G. Catalano, and A.~Bicchi, \emph{Soft Robots}.\hskip 1em
  plus 0.5em minus 0.4em\relax Berlin, Heidelberg: Springer, 2020, pp. 1--15.

\bibitem{zha20}
C.~Zhang, W.~Zou, L.~Ma, and Z.~Wang, ``Biologically inspired jumping robots: A
  comprehensive review,'' \emph{Robotics and Autonomous Systems}, vol. 124, p.
  103362, 2020.

\bibitem{bad22}
A.~Badri-Spr{\"o}witz, A.~Aghamaleki~Sarvestani, M.~Sitti, and M.~A. Daley,
  ``Birdbot achieves energy-efficient gait with minimal control using
  avian-inspired leg clutching,'' \emph{Science Robotics}, vol.~7, no.~64,
  2022.

\bibitem{frund2023bipedal}
K.~Fr{\"u}nd, F.~Beck, A.~Shu, F.~Loeffl, and J.~Lee, ``Bipedal running:
  Bioinspired fundamentals for versatile humanoid robot locomotion,'' in
  \emph{2023 IEEE-RAS 22nd International Conference on Humanoid Robots
  (Humanoids)}.\hskip 1em plus 0.5em minus 0.4em\relax IEEE, 2023, pp. 1--8.

\bibitem{rut08}
S.~Rutishauser, A.~Sprowitz, L.~Righetti, and A.~J. Ijspeert, ``Passive
  compliant quadruped robot using central pattern generators for locomotion
  control,'' in \emph{2008 2nd IEEE RAS \& EMBS International Conference on
  Biomedical Robotics and Biomechatronics}.\hskip 1em plus 0.5em minus
  0.4em\relax IEEE, 2008, pp. 710--715.

\bibitem{che19}
J.~Chen, Z.~Liang, Y.~Zhu, C.~Liu, L.~Zhang, L.~Hao, and J.~Zhao, ``Towards the
  exploitation of physical compliance in segmented and electrically actuated
  robotic legs: A review focused on elastic mechanisms,'' \emph{Sensors},
  vol.~19, no.~24, p. 5351, 2019.

\bibitem{cal23}
D.~Calzolari, C.~Della~Santina, A.~M. Giordano, A.~Schmidt, and
  A.~Albu-Sch{\"a}ffer, ``Embodying quasi-passive modal trotting and pronking
  in a sagittal elastic quadruped,'' \emph{IEEE Robotics and Automation
  Letters}, vol.~8, no.~4, pp. 2285--2292, 2023.

\bibitem{ijs23}
A.~J. Ijspeert and M.~A. Daley, ``Integration of feedforward and feedback
  control in the neuromechanics of vertebrate locomotion: a review of
  experimental, simulation and robotic studies,'' \emph{Journal of Experimental
  Biology}, vol. 226, no.~15, p. jeb245784, 2023.

\bibitem{McGeer1990}
T.~McGeer, ``Passive bipedal running,'' \emph{Proceedings of the Royal Society
  of London. B. Biological Sciences}, vol. 240, no. 1297, pp. 107--134, 1990.

\bibitem{Blickh1989}
R.~Blickhan, ``The spring-mass model for running and hopping,'' \emph{Journal
  of Biomechanics}, vol.~22, pp. 1217--1227, 1989.

\bibitem{poulakakis2009spring}
I.~Poulakakis and J.~W. Grizzle, ``The spring loaded inverted pendulum as the
  hybrid zero dynamics of an asymmetric hopper,'' \emph{IEEE Transactions on
  Automatic Control}, vol.~54, no.~8, pp. 1779--1793, 2009.

\bibitem{Geyer2006}
H.~Geyer, A.~Seyfarth, and R.~Blickhan, ``Compliant leg behavior explains basic
  dynamics of walking and running,'' \emph{Proceedings of the Royal Society B},
  vol. 273, pp. 2861--2867, nov 2006.

\bibitem{Hutter2012}
M.~Hutter, C.~D. Remy, M.~A. Hoepflinger, and R.~Siegwart, ``High compliant
  series elastic actuation for the robotic leg scarleth,'' in \emph{Field
  Robotics}.\hskip 1em plus 0.5em minus 0.4em\relax World Scientific, 2012, pp.
  507--514.

\bibitem{ruppert2022nature}
F.~Ruppert and A.~Badri-Spr{\"o}witz, ``Learning plastic matching of robot
  dynamics in closed-loop central pattern generators,'' \emph{Nature Machine
  Intelligence}, vol.~4, no.~7, pp. 652--660, 2022.

\bibitem{stella2025synergy}
F.~Stella, M.~M. Achkar, C.~Della~Santina, and J.~Hughes, ``Synergy-based
  robotic quadruped leveraging passivity for natural intelligence and
  behavioural diversity,'' \emph{Nature Machine Intelligence}, pp. 1--14, 2025.

\bibitem{remy16}
W.~Xi, Y.~Yesilevskiy, and C.~D. Remy, ``Selecting gaits for economical
  locomotion of legged robots,'' \emph{The International Journal of Robotics
  Research}, vol.~35, no.~9, pp. 1140--1154, 2016.

\bibitem{raff2022generating}
M.~Raff, N.~Rosa, and C.~D. Remy, ``Generating families of optimally actuated
  gaits from a legged system's energetically conservative dynamics,'' in
  \emph{2022 IEEE/RSJ International Conference on Intelligent Robots and
  Systems (IROS)}.\hskip 1em plus 0.5em minus 0.4em\relax IEEE, 2022, pp.
  8866--8872.

\bibitem{ijspeert2008central}
A.~J. Ijspeert, ``Central pattern generators for locomotion control in animals
  and robots: a review,'' \emph{Neural networks}, vol.~21, no.~4, pp. 642--653,
  2008.

\bibitem{ram23}
P.~Ramdya and A.~J. Ijspeert, ``The neuromechanics of animal locomotion: From
  biology to robotics and back,'' \emph{Science Robotics}, vol.~8, no.~78,
  2023.

\bibitem{alb20}
A.~Albu-Sch{\"a}ffer and C.~Della~Santina, ``A review on nonlinear modes in
  conservative mechanical systems,'' \emph{Annual Reviews in Control}, vol.~50,
  pp. 49--71, 2020.

\bibitem{sac22}
A.~Sachtler and A.~Albu-Sch{\"a}ffer, ``Strict modes everywhere--bringing order
  into dynamics of mechanical systems by a potential compatible with the
  geodesic flow,'' \emph{IEEE Robotics and Automation Letters}, vol.~7, no.~2,
  pp. 2337--2344, 2022.

\bibitem{bje22}
F.~Bjelonic, A.~Sachtler, A.~Albu-Sch{\"{a}}ffer, and C.~{Della Santina},
  ``{Experimental Closed-Loop Excitation of Nonlinear Normal Modes on an
  Elastic Industrial Robot},'' \emph{IEEE Robotics and Automation Letters},
  vol.~7, no.~2, pp. 1689--1696, apr 2022.

\bibitem{alb21}
A.~Albu-Sch{\"a}ffer, D.~Lakatos, and S.~Stramigioli, ``Strict nonlinear normal
  modes of systems characterized by scalar functions on riemannian manifolds,''
  \emph{IEEE Robotics and Automation Letters}, vol.~6, no.~2, pp. 1910--1917,
  2021.

\bibitem{raffin2023simple}
A.~Raffin, O.~Sigaud, J.~Kober, A.~Albu-Sch{\"a}ffer, J.~Silv{\'{e}}rio, and
  F.~Stulp, ``An open-loop baseline for reinforcement learning locomotion
  tasks,'' \emph{Reinforcement Learning Journal}, vol.~1, 2024.

\bibitem{farley1993jeb}
C.~T. Farley, J.~Glasheen, and T.~A. McMahon, ``Running springs: Speed and
  animal size,'' \emph{Journal of Experimental Biology}, vol. 185, no.~1, pp.
  71--86, 1993.

\bibitem{kim2013soft}
S.~Kim, C.~Laschi, and B.~Trimmer, ``Soft robotics: a bioinspired evolution in
  robotics,'' \emph{Trends in biotechnology}, vol.~31, no.~5, pp. 287--294,
  2013.

\bibitem{lak14}
D.~Lakatos, F.~Petit, and A.~Albu-Sch{\"a}ffer, ``Nonlinear oscillations for
  cyclic movements in human and robotic arms,'' \emph{IEEE Transactions on
  Robotics}, vol.~30, no.~4, pp. 865--879, 2014.

\bibitem{manipulandum}
P.~Stratmann, A.~Schmidt, H.~H{\"o}ppner, P.~van~der Smagt, T.~Meindl, D.~W.
  Franklin, and A.~Albu-Sch{\"a}ffer, ``Human short-latency reflexes show
  precise short-term gain adaptation after prior motion,'' \emph{Journal of
  Neurophysiology}, vol. 132, no.~6, pp. 1680--1692, 2024.

\bibitem{sch21}
A.~Schmidt, B.~Feldotto, T.~Gumpert, D.~Seidel, A.~Albu-Sch{\"a}ffer, and
  P.~Stratmann, ``Adapting highly-dynamic compliant movements to changing
  environments: A benchmark comparison of reflex-vs. cpg-based control
  strategies,'' \emph{Frontiers in Neurorobotics}, vol.~15, p. 762431, 2021.

\bibitem{joh22}
T.~A. Johnson, W.~J. Gordon-Evans, B.~D.~X. Lascelles, and M.~G. Conzemius,
  ``Determination of the center of mass in a heterogeneous population of
  dogs,'' \emph{Plos one}, vol.~17, no.~4, p. e0267361, 2022.

\bibitem{Zink2020dogposture}
C.~Zink and M.~R. Schlehr, ``Working dog structure: evaluation and relationship
  to function,'' \emph{Frontiers in Veterinary Science}, vol.~7, p. 559055,
  2020.

\bibitem{raibert1984hopping}
M.~H. Raibert, ``Hopping in legged systems—modeling and simulation for the
  two-dimensional one-legged case,'' \emph{IEEE Transactions on Systems, Man,
  and Cybernetics}, no.~3, pp. 451--463, 1984.

\bibitem{hereid2014dynamic}
A.~Hereid, S.~Kolathaya, M.~S. Jones, J.~Van~Why, J.~W. Hurst, and A.~D. Ames,
  ``Dynamic multi-domain bipedal walking with atrias through slip based
  human-inspired control,'' in \emph{Proceedings of the 17th international
  conference on Hybrid systems: computation and control}, 2014, pp. 263--272.

\bibitem{reher2019dynamic}
J.~Reher, W.-L. Ma, and A.~D. Ames, ``Dynamic walking with compliance on a
  cassie bipedal robot,'' in \emph{2019 18th European Control Conference
  (ECC)}.\hskip 1em plus 0.5em minus 0.4em\relax IEEE, 2019, pp. 2589--2595.

\bibitem{Calzolari2024tofix}
D.~Calzolari, C.~Della~Santina, and A.~Albu-Sch{\"a}ffer, ``Exciting families
  of passive gaits in an elastic quadruped via natural motion manifold
  control,'' \emph{The International Journal of Robotics Research}, vol.~45,
  no.~2, pp. 233--258, 2026.

\bibitem{Gan2014}
Z.~Gan and C.~D. Remy, ``A passive dynamic quadruped that moves in a large
  variety of gaits,'' in \emph{2014 IEEE/RSJ International Conference on
  Intelligent Robots and Systems}, 2014, pp. 4876--4881.

\bibitem{Sachtler2024swingup}
A.~Sachtler, D.~Calzolari, M.~Raff, A.~Schmidt, Y.~P. Wotte, C.~D. Santina,
  C.~D. Remy, and A.~Albu-Schäffer, ``Swing-up of a weakly actuated double
  pendulum via nonlinear normal modes,'' in \emph{European Control Conference},
  4 2024.

\bibitem{ehlert2026locomotion}
T.~Ehlert, A.~Sachtler, A.~Schmidt, D.~Calzolari, and A.~Albu-Sch{\"a}ffer,
  ``Locomotion of an elastic snake robot via natural dynamics,'' \emph{IEEE
  Robotics and Automation Letters}, 2026.

\bibitem{biewener2006patterns}
A.~A. Biewener, ``Patterns of mechanical energy change in tetrapod gait:
  pendula, springs and work,'' \emph{Journal of Experimental Zoology Part A:
  Comparative Experimental Biology}, vol. 305, no.~11, pp. 899--911, 2006.

\bibitem{sesselmann2021embedding}
A.~Sesselmann, F.~Loeffl, C.~Della~Santina, M.~A. Roa, and
  A.~Albu-Sch{\"a}ffer, ``Embedding a nonlinear strict oscillatory mode into a
  segmented leg,'' in \emph{2021 IEEE/RSJ International Conference on
  Intelligent Robots and Systems (IROS)}.\hskip 1em plus 0.5em minus
  0.4em\relax IEEE, 2021, pp. 1370--1377.

\bibitem{srinivasanRuina2006computer}
M.~Srinivasan and A.~Ruina, ``Computer optimization of a minimal biped model
  discovers walking and running,'' \emph{Nature}, vol. 439, no. 7072, pp.
  72--75, 2006.

\bibitem{ijspeert2013dynamical}
A.~J. Ijspeert, J.~Nakanishi, H.~Hoffmann, P.~Pastor, and S.~Schaal,
  ``Dynamical movement primitives: learning attractor models for motor
  behaviors,'' \emph{Neural computation}, vol.~25, no.~2, pp. 328--373, 2013.

\bibitem{pustina2024nonlinear}
P.~Pustina, D.~Calzolari, A.~Albu-Sch{\"a}ffer, A.~De~Luca, and
  C.~Della~Santina, ``Nonlinear modes as a tool for comparing the mathematical
  structure of dynamic models of soft robots,'' in \emph{2024 IEEE 7th
  International Conference on Soft Robotics (RoboSoft)}.\hskip 1em plus 0.5em
  minus 0.4em\relax IEEE, 2024, pp. 779--785.

\bibitem{gaba2026comprehensive}
S.~Gaba, K.~Rana, S.~Sai, V.~Chamola, and D.~Niyato, ``A comprehensive review
  of generative physical artificial intelligence,'' \emph{IEEE Internet of
  Things Journal}, 2026.

\bibitem{whitebert}
D.~{Lakatos}, K.~{Ploeger}, F.~{Loeffl}, D.~{Seidel}, F.~{Schmidt},
  T.~{Gumpert}, F.~{John}, T.~{Bertram}, and A.~{Albu-Schäffer}, ``Dynamic
  locomotion gaits of a compliantly actuated quadruped with slip-like
  articulated legs embodied in the mechanical design,'' \emph{IEEE Robotics and
  Automation Letters}, vol.~3, no.~4, pp. 3908--3915, Oct 2018.

\bibitem{morey1992dogsize}
D.~F. Morey, ``Size, shape and development in the evolution of the domestic
  dog,'' \emph{Journal of Archaeological Science}, vol.~19, no.~2, pp.
  181--204, 1992.

\bibitem{Chen2000}
S.-F. Chen and I.~Kao, ``Conservative congruence transformation for joint and
  cartesian stiffness matrices of robotic hands and fingers,'' \emph{The
  International Journal of Robotics Research}, vol.~19, no.~9, pp. 835--847,
  2000.

\bibitem{schmidt2026links}
F.~Schmidt, J.~Nix, M.~M{\"u}hlbauer, J.~Cremer, M.~Chalon, T.~Bachmann, and
  A.~Raffin, ``Links and nodes: Middleware for distributed real-time robotic
  systems,'' \emph{Journal of Open Source Software}, vol.~11, no. 124, p.
  10777, 2026.

\bibitem{ROSENBERG1966}
R.~Rosenberg, ``On nonlinear vibrations of systems with many degrees of
  freedom,'' \emph{Advances in applied mechanics}, vol.~9, pp. 155--242, 1966.

\bibitem{stra16}
P.~Stratmann, D.~Lakatos, M.~C. {\"O}zparpucu, and A.~Albu-Sch{\"a}ffer,
  ``Legged elastic multibody systems: adjusting limit cycles to
  close-to-optimal energy efficiency,'' \emph{IEEE Robotics and Automation
  Letters}, vol.~2, no.~2, pp. 436--443, 2016.

\bibitem{wandinger2024tele}
D.~Wandinger, A.~Schmidt, A.~Raffin, A.~Albu-Sch{\"a}ffer, and M.~Keppler,
  ``Tele-running: Trajectory generation for monopod robots by teleoperation,''
  in \emph{2024 IEEE International Conference on Systems, Man, and Cybernetics
  (SMC)}.\hskip 1em plus 0.5em minus 0.4em\relax IEEE, 2024, pp. 273--278.

\bibitem{seidel2024toward}
D.~Seidel, A.~Schmidt, X.~Luo, A.~Raffin, L.~Mayershofer, T.~Ehlert,
  D.~Calzolari, M.~Hermann, T.~Gumpert, F.~L{\"o}ffl, E.~{den Exter},
  A.~K{\"o}pken, R.~Luz, A.~Bauer, N.~Batti, F.~Lay, A.~Manaparampil,
  A.~{Albu-Sch{\"a}ffer}, D.~Leidner, P.~Schmau, T.~Kr{\"u}ger, and N.~Y. Lii,
  ``Toward space exploration on legs: Iss-to-earth teleoperation experiments
  with a quadruped robot,'' in \emph{2024 IEEE Conference on
  Telepresence}.\hskip 1em plus 0.5em minus 0.4em\relax IEEE, 2024, pp. 10--15.

\bibitem{stra16b}
P.~Stratmann, D.~Lakatos, and A.~Albu-Sch{\"a}ffer, ``Neuromodulation and
  synaptic plasticity for the control of fast periodic movement: energy
  efficiency in coupled compliant joints via pca,'' \emph{Frontiers in
  neurorobotics}, vol.~10, p.~2, 2016.

\bibitem{Hutter2013efficient}
M.~Hutter, C.~D. Remy, M.~A. Hoepflinger, and R.~Siegwart, ``Efficient and
  versatile locomotion with highly compliant legs,'' \emph{IEEE/ASME
  Transactions on Mechatronics}, vol.~18, pp. 449--458, 2013.

\bibitem{hansen2001completely}
N.~Hansen and A.~Ostermeier, ``Completely derandomized self-adaptation in
  evolution strategies,'' \emph{Evolutionary computation}, vol.~9, no.~2, pp.
  159--195, 2001.

\bibitem{takuya2019optuna}
T.~Akiba, S.~Sano, T.~Yanase, T.~Ohta, and M.~Koyama, ``Optuna: A
  next-generation hyperparameter optimization framework,'' in \emph{Proceedings
  of the 25th ACM SIGKDD International Conference on Knowledge Discovery \&amp;
  Data Mining}, ser. KDD '19.\hskip 1em plus 0.5em minus 0.4em\relax New York,
  NY, USA: Association for Computing Machinery, 2019, p. 2623–2631.

\bibitem{ebert-figshare}
\BIBentryALTinterwordspacing
A.~Schmidt, ``{eBert} data and analysis code,'' 2026-08-31. [Online].
  Available: \url{https://doi.org/10.6084/m9.figshare.29514518}
\BIBentrySTDinterwordspacing

\bibitem{ebert-git}
\BIBentryALTinterwordspacing
D.~Calzolari, A.~Sachtler, A.~Raffin, and A.~Schmidt, ``{eBert} modes,'' 9
  2026. [Online]. Available: \url{https://github.com/anschm189/ebert_modes.git}
\BIBentrySTDinterwordspacing

\bibitem{pol22}
M.~J. Pollayil, C.~Della~Santina, G.~Mesesan, J.~Englsberger, D.~Seidel,
  M.~Garabini, C.~Ott, A.~Bicchi, and A.~Albu-Schaffer, ``Planning natural
  locomotion for articulated soft quadrupeds,'' in \emph{2022 International
  Conference on Robotics and Automation (ICRA)}.\hskip 1em plus 0.5em minus
  0.4em\relax IEEE, 2022, pp. 6593--6599.

\bibitem{Michael2024dogCOM}
H.~E. Michael, C.~M. McGowan, and H.~K. Hyytiäinen, ``Posture and postural
  dysfunction in dogs: Implications for veterinary physiotherapy,'' \emph{The
  Veterinary Journal}, vol. 305, p. 106107, 2024.

\end{thebibliography}

\appendices
\setcounter{figure}{0}
\setcounter{table}{0}
\renewcommand{\thefigure}{S\arabic{figure}}
\renewcommand{\thetable}{S\arabic{table}}
\setcounter{equation}{0}
\renewcommand{\theequation}{S\arabic{equation}}

\section{Supplementary Methods}
\addcontentsline{toc}{section}{Supplementary Methods}
\label{sec:supp}

\subsection{Nonlinear modal analysis}
\begin{figure}
    \centering
    \includegraphics[width=\linewidth]{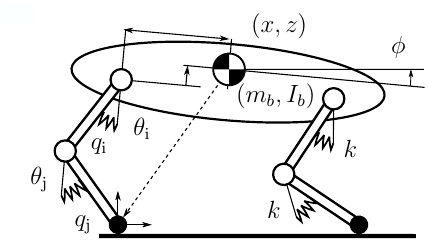}
    \caption{\textbf{Sagittal view of the conservative model representing \ebert.} This serially elastically actuated quadruped model was used to compute the modal oscillations.}
    \label{fig:model}
\end{figure}

\addcontentsline{toc}{subsection}{Nonlinear modal analysis}
\label{sec:app:nnm}
The numerical continuation method used to compute the NNMs is based on established algorithms from prior work \cite{bje22, alb21, alb20}. This section details the adaptation and application of this method to the specific kinematic and dynamic model of the \ebert quadruped.
To identify the nonlinear modes of the quadruped robot, we utilize a conservative model based solely on the interaction of multi-body dynamics with a potential field $V$ comprising gravitational and elastic parts (Fig.~\ref{fig:model}).
The equations of motion of a serially elastically actuated quadruped are
\begin{align}
    \small
    \label{eq:quadruped_joint_dynamics}
    \M(\z)\ddz + \bs{c}(\z,\dz) + \bm{g}(\z) &= \bm{S}^T \bm{K}(\thm - \q) + \bm{J}_{c}^\tr(\bm{q})\bm{\lambda} \\
    \label{eq:quadruped_motor_dynamics}
    \bm{B}\ddthm + \bm{K}(\bm{\theta}-\bm{q}) &= \tauv \;,
\end{align}
where the coordinates $\bs{z}=(\x, \q) \inR{n}$ comprise the trunk position and orientation $\x\inR{6}$ and the leg link positions $\q \inR{n_j}$.
The symbol $\M$ denotes the inertia matrix, $\cc$ collects the Coriolis and centrifugal forces, while $\gv$ are the gravitational forces.
The motor positions are denoted by $\bm{\theta} \inR{n_j}$, the respective motor torques by $\bm{\tau}$, and the projected motor inertia by $\bm{B}$.
The matrix $\bm{K}$ is the joint spring stiffness, while $\bm{S}$ is a selection matrix.
The contact forces $\bm{\lambda}\inR{n_\lambda}$ are projected by the contact Jacobian $\bm{J}_c\inR{n \times n_\lambda}$.

To find the modes of the mechanical system, we remove the motor dynamics by fixing $\thm$ to a constant value (i.e., the rest configuration of the springs).
Then, the system is rewritten in a minimal set of coordinates. When the four feet are fixed to the ground, the robot is left with 6 DOF.
Therefore, $\x$ is taken as a minimal coordinate.
By partitioning $\z$ into the independent and dependent variables, $\x$ and $\q$, respectively, and utilizing the Jacobian of the constraint equation for the feet, we obtain
\begin{equation}
\begin{split}
     [\bm{J}_{c,x} \ \bm{J}_{c,q}]
     \begin{bmatrix}
     \dx \\ \dq
     \end{bmatrix}
     = \bm{0} \ \rightarrow \ \dq = -\bm{J}_{c,q}^{-1} \bm{J}_{c,x} \dx \ .
\end{split}
\end{equation}
We can then express the generalized velocities $\dz = (\dx, \dq)$ in terms of the independent variables via a transformation $\bm{T}(\x)$ such that $\dz = \bm{T}(\x)\dx$ \cite{pol22}. Additionally, the inverse kinematics $\q = f_\mathrm{ik}{}(\x)$ is assumed to be available.

The system can now be written as
\begin{equation}\label{eq:eom_minimal}
   \M_x(\x)\ddx + \cc_x(\x,\dx) + \gv_x(\x) + \parX{{}^\tr V(\x)}{\x} = \bm{0} \;,
\end{equation}
where $\bm{M}_x(\bm{x}) = \bm{T}^\tr \bm{M} \bm{T}$, $\cc_x(\x,\dx)$ are the Coriolis/centrifugal forces and $\gv_x(\x)$ the gravity forces in this compact representation. The total mechanical energy is $E = \half\dx^\tr\M_x(\x)\dx + V(\x)$.
\newline

\paragraph*{Linear modes computation}
The linearization around a stable equilibrium $\xeq$ yields
\begin{align}
    \label{eq:linearized_qI_sys}
    \M_x(\xeq)\ddot{\tilde{\x}} + \K_x(\xeq)\tilde{\x} = \bm{0} \;,
\end{align}
where $\tilde{\x} = \x - \xeq$ is the deviation from the equilibrium and $\bm{K}_x(\bm{x})  \approx \bm{T}^\tr (\bm{S}^\tr \bm{K} \bm{S} + \frac{\partial \bm{g}} {\partial \bm{z}}) \bm{T}$.
For each eigenvalue $\lambda_i$ and eigenvector $\bm{v}_i$ pair, Equation~\eqref{eq:linearized_qI_sys} leads to solutions of the form
\begin{equation}
    \tilde{\x}(t) = \bm{v}_i  A  \sin(\omega_i t + \varphi)
\end{equation}
where $\omega_i = \sqrt{\lambda_i}$ are the oscillation frequencies, while the amplitude $A$ and the phase $\varphi$ depend on the initial conditions.
For example, the linearized mode vectors in joint space of tilted position, which was determined for \ebert to develop locomotion following the methods from \hyperref[sec:app:grid]{Determining Initial Configuration for Locomotion}, are summarized in Table~\ref{tab:method:foot-pair}.

\begin{table}
\centering
\caption{\textbf{Linearized mode vectors for the tilted walking configuration of \ebert in joint space.} The color code points out respective similarities of the shoulder (s), hip (h), and knee (k) joints, defining which legs should be moved simultaneously as shown in Fig.~\ref{fig:loco-modes}C.}
\label{tab:method:foot-pair}
\begin{tabular}{lc|cccccc}
                       & & $\mathcal{M}_1$                      & $\mathcal{M}_2$ & $\mathcal{M}_3$ & $\mathcal{M}_4$ & $\mathcal{M}_5$ & $\mathcal{M}_6$ \\
                       \hline
                       &s & \textcolor{magenta}{+0.75} & \textcolor{cyan}{-0.07}    & \textcolor{magenta}{\underline{+0.97}}
                          & \textcolor{black}{-0.38} & \textcolor{cyan}{-0.08} & \textcolor{magenta}{-1.08} \\
                       &h & \textcolor{magenta}{+1.31} & \textcolor{cyan}{+4.30}    & \textcolor{black}{-0.34}
                          & \textcolor{cyan}{\underline{+7.22}}  & \textcolor{cyan}{+0.58}  & \textcolor{magenta}{+0.95}  \\
\multirow{-3}{*}{\textbf{FR}}
                       &k & \textcolor{magenta}{-1.10} & \textcolor{cyan}{+1.31}    & \textcolor{black}{-0.65}
                         & \textcolor{cyan}{\underline{-8.74}} & \textcolor{cyan}{-2.47} & \textcolor{magenta}{-0.94} \\
                        \hline
                       &s & \textcolor{cyan}{-0.75}    & \textcolor{cyan}{-0.07}    & \textcolor{cyan}{\underline{-0.97}}
                         & \textcolor{black}{-0.40} & \textcolor{cyan}{-0.07} & \textcolor{cyan}{+1.08}  \\
                       &h & \textcolor{cyan}{-1.31}    & \textcolor{cyan}{+4.30}    & \textcolor{black}{+0.34}
                         & \textcolor{cyan}{\underline{+7.23}}  & \textcolor{cyan}{+0.57}  & \textcolor{cyan}{-0.95} \\
\multirow{-3}{*}{\textbf{FL}}
                       &k & \textcolor{cyan}{+1.10}    & \textcolor{cyan}{+1.31}    & \textcolor{black}{+0.65}
                         & \textcolor{cyan}{\underline{-8.74}} & \textcolor{cyan}{-2.46} & \textcolor{cyan}{+0.95}  \\
                        \hline
                       &s & \textcolor{magenta}{+1.14} & \textcolor{magenta}{-0.02} & \textcolor{cyan}{\underline{-0.79}}
                         & \textcolor{black}{+0.13}  & \textcolor{magenta}{-0.06} & \textcolor{magenta}{-1.00} \\
                       &h & \textcolor{magenta}{+0.82} & \textcolor{magenta}{+1.36} & \textcolor{black}{-0.25}
                         & \textcolor{cyan}{\underline{+5.30}}  & \textcolor{magenta}{-3.26} & \textcolor{magenta}{+1.12}  \\
\multirow{-3}{*}{\textbf{HR}}
                       &k & \textcolor{magenta}{-0.64} & \textcolor{magenta}{+4.07} & \textcolor{black}{-0.46}
                         & \textcolor{cyan}{\underline{-6.60}} & \textcolor{magenta}{+1.27}  & \textcolor{magenta}{-1.08} \\
                         \hline
                       &s & \textcolor{cyan}{-1.14}    & \textcolor{magenta}{-0.02} & \textcolor{magenta}{\underline{+0.79}}
                         & \textcolor{black}{+0.12}  & \textcolor{magenta}{-0.05} & \textcolor{cyan}{+1.00}  \\
                       &h & \textcolor{cyan}{-0.82}    & \textcolor{magenta}{+1.36} & \textcolor{black}{+0.25}
                         & \textcolor{cyan}{\underline{+5.31}}  & \textcolor{magenta}{-3.27} & \textcolor{cyan}{-1.11} \\
\multirow{-3}{*}{\textbf{HL}}
                       &k & \textcolor{cyan}{+0.64}    & \textcolor{magenta}{+4.07} & \textcolor{black}{+0.46}
                         & \textcolor{cyan}{\underline{-6.60}} & \textcolor{magenta}{+1.28}  & \textcolor{cyan}{+1.07}
        \end{tabular}
\end{table}

\paragraph*{Nonlinear modes computation}
To compute an Eigenmanifold, we start by investigating the behavior of the nonlinear system Equation~\eqref{eq:eom_minimal} in a neighborhood of the equilibrium.
If we initialize the system with a small deviation $\x(0) = \xeq + \epsilon \vv_\ii$ at $\dx(0) = \boldsymbol{0}$, system Equation~\eqref{eq:eom_minimal} behaves like a harmonic oscillator with frequency $\omega_\ii$ and along $\vv_\ii$.
These periodic orbits will have exactly two turning points where the velocity is zero. We call this type of orbit \emph{brake orbit}.

As we increase the energy, the nonlinearity will prevent these periodic orbits from existing in the Eigenplanes spawned by the eigenvectors.
We can, however, recover a periodic motion by adjusting the turning points.
This is a numerical continuation problem starting from the known linear mode, with the constraint that the periodic orbits must have exactly two turning points.
By iteratively increasing the energy and adjusting the turning points, this process creates a continuous family of brake orbits starting from the $i$-th linear mode.
We call this collection of orbits the \emph{$i$}-th nonlinear normal mode.
Since this process can be performed for each eigenvector, a system with $n$ DOFs is expected to have at least $n$ Eigenmanifolds.

The turning points $\xturn$ of the nonlinear mode can now be described by a function $\xturn = \GG_{i\pm}(E)$, which we choose to parameterize by the energy $E$.
By initializing the system to either $\GG_{i+}(E)$ or $\GG_{i-}(E)$ with zero velocity, we obtain a periodic orbit oscillating between the two turning points. Formally
\begin{alignat}{4}
    \x(0) &= \GG_{i\pm}(E) &\quad \dx(0) &= \boldsymbol{0} \nonumber\\
    \x(t) &= \x(t + T_i(E)) &\quad \dx(t) &= \dx(t + T_i(E)) \;.\nonumber
\end{alignat}
We call the functions $\GG_{i\pm}(E)$ the \emph{generators} of the $i$-th mode \cite{alb20}.
In this sense, they can be interpreted as the nonlinear counterparts of eigenvectors.

\begin{figure*}
    \includegraphics[width=0.52\linewidth]{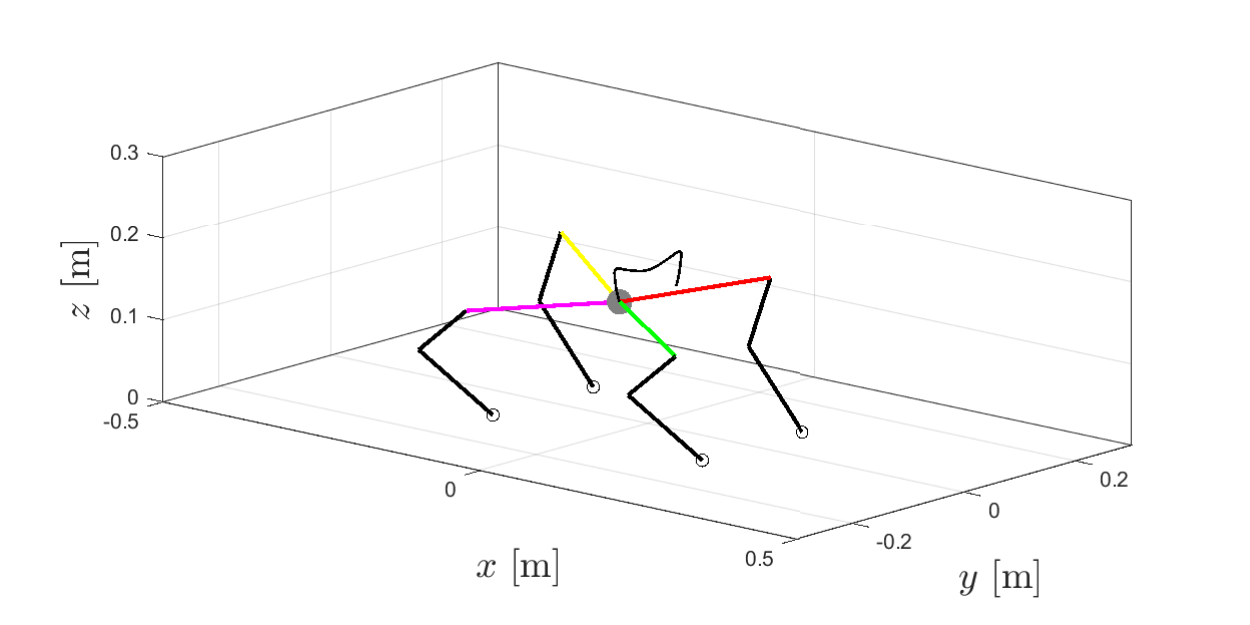}
    \hfill
    \includegraphics[width=0.52\linewidth]{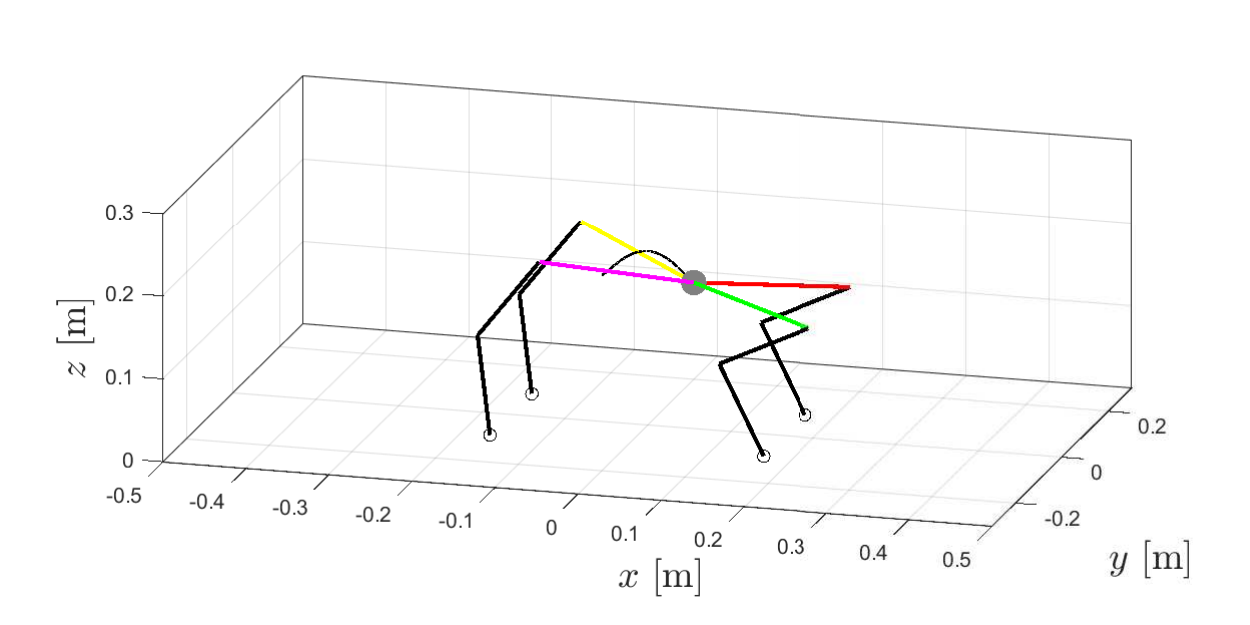}
    \caption{\textbf{Visualization of \ebert's first two nonlinear modal oscillations}. To compute them numerical continuation was applied.}
    \label{fig:modetool_sim_vis}
\end{figure*}

Finally, when collecting all trajectories of the $i$-th NNM for each energy level $E$, we get a two-dimensional submanifold $\mathcal{M}_\ii \subset \mathbb{R}^{2n}$ of the state space, i.e., the Eigenmanifold
\begin{equation*}\label{eq:eigenmanifold}
    \begingroup\setlength\arraycolsep{1pt}
    \mathcal{M}_i = \left\{ \begin{bmatrix}\x(t) \\ \dx(t)\end{bmatrix} \in \mathbb{R}^{2n} \right. \left|\begin{array}{lcl}\x(0) &=& \GG_{i\pm}(E),\\\dx(0) &=& \boldsymbol{0},\end{array}\hfill \begin{array}{lcl}E & \in & [0, E_{\mathrm{max}}),\\t &\in& [0, T_i(E)).\end{array} \right\},
    \endgroup
\end{equation*}
where $[\x(t), \dx(t)]^\tr$ is a solution to~Equation~\eqref{eq:quadruped_joint_dynamics}.
Fig.~\ref{fig:modetool_sim_vis} reports oscillations on two distinct Eigenmanifolds of \ebert.

\subsection{Quantifying Nonlinearity of the Modes}
\addcontentsline{toc}{subsection}{Quantifying Nonlinearity of the Modes}
\label{sec:app:quantify-nonlinearity}
To verify that the modes are truly nonlinear, i.e., that linear theory or linearization is not sufficient to describe and analyze the system, we quantify the nonlinearity by measuring the error introduced by a linear approximation.

Each point on the Eigenmanifold is a state $[\x, \dx]^\tr \in \mathcal{M}_i$ of the system that belongs to the $i$-th mode.
As outlined above, we use a numerical continuation algorithm that generates a data-driven approximation of the Eigenmanifold $\mathcal{M}_i$:
by sweeping through energy and collecting all time-points along the periodic orbits, we obtain a point cloud representation of $\mathcal{M}_i$.

The main idea to quantify the nonlinearity is the following:
if the mode $\mathcal{M}_i$ is linear, all points will lie on a two-dimensional plane in the state space;
and if the mode is nonlinear, the Eigenmanifold will be curved and require a volume of higher dimension to contain all points.

\begin{figure*}
\centering
\includegraphics[width=\textwidth]{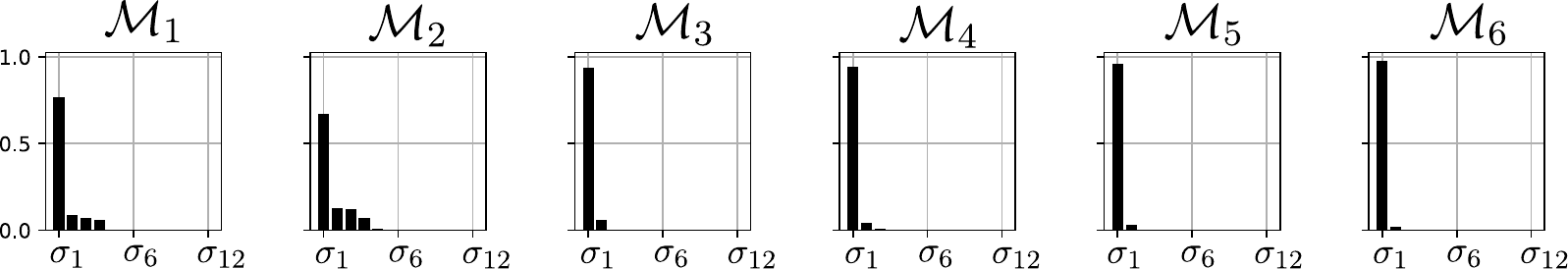}
\caption{\textbf{Singular values of reconstructing the modes.}}
\label{fig:app:svdvals}
\end{figure*}

To quantify this, we apply a Principal Component Analysis (PCA) to the point cloud of the Eigenmanifold $\mathcal{M}_i$.
Therefore, we collect all points into a data matrix $\bm{X}\in\mathbb{R}^{m\times 12}$ \begin{equation}
    \bm{X}(\mathcal{M}_i) = \begin{bmatrix}
    \x_1^\tr & \dx_1^\tr \\
    \vdots & \vdots \\
    \x_m^\tr & \dx_m^\tr
    \end{bmatrix} \text{ for } [\x_i, \dx_i]^\tr \in \mathcal{M}_i,
\end{equation}
where $m$ is the number of points in the point cloud.
Then we compute the singular value decomposition \begin{equation}
    \bm{X}(\mathcal{M}_i) - \bm{\bar{X}}(\mathcal{M}_i) = \bm{U} \bm{\Sigma} \bm{V}^\tr,
\end{equation}
where $\bm{\bar{X}}(\mathcal{M}_i)$ is the column-wise mean of the point cloud and $\bm{\Sigma}$ contains the twelve singular values $\bar{\sigma}_i$ on its diagonal.
We normalize the singular values to $\sum \sigma_i = 1$ by \begin{equation}
    \sigma_i = \frac{\bar{\sigma}_i}{\sum_{j} \bar{\sigma}_j} \;,
\end{equation}
and report the results in Fig.~\ref{fig:app:svdvals}.
We see that the first two modes each have four singular values that are significantly larger than zero.
Hence, for those modes we need at least a four-dimensional volume to contain all points of the Eigenmanifold and the system cannot be approximated by a plane, i.e., a linear mode.

\subsection{Robotic Implementation}
\addcontentsline{toc}{subsection}{Robotic Implementation}
\label{sec:app:robot}
Following the hardware implementations of the previous \textit{bert} versions \cite{whitebert}, \ebert was designed to be low-cost with off-the-shelf components and rapid prototyping methods, building on the gained knowledge from the earlier quadruped versions.

\paragraph*{Body fabrication}
All body parts were manufactured from polyamide using direct laser sintering to provide the load-carrying shell that protects the inner components and forms the robot's physical appearance (Fig.~\ref{fig:bert-modes}A).
The onboard computer and most electronics were housed in the base body of \ebert (Fig.~\ref{fig:hardware}A), while the shoulder modules included the three individual SEA units for each leg (Fig.~\ref{fig:hardware}C), as well as the modules for the link sensors. Consequently, the main mass of \ebert was concentrated in the body's trunk, while the legs were comparatively massless as initially derived from the work from Gan et al \cite{Gan2014}.
The biarticular coupling of the hip and knee joints was implemented via synchronous belt drives (Fig.~\ref{fig:hardware}B).

\paragraph*{Actuation}
The actuator in each SEA was a servo drive based on the commercially available Savöx SB-2290SG. It incorporated a brushless motor and a four-stage steel gearbox with a 321:1 gear ratio to transmit the power to the output shaft (Fig.~\ref{fig:hardware}C).
The electronics were replaced in-house to provide necessary telemetry data and control accuracy for robot control.
The custom PCBs incorporated a discrete 3-phase GaN-based power stage, a 32-bit STM microcontroller with a \SI{400}{\mega \hertz} clock rate, and a contactless 14-bit on-axis magnetic position sensor based on the Hall effect.
The brushless motor was internally operated at \SI{12}{\volt} with a
\SI{250}{\kilo \hertz} current controller.
The communication interface to the control host was implemented using DS402 EtherCAT with a communication rate of \SI{1}{\kilo \hertz}.
The peak forward torque was \SI{2,9}{\newton \meter}. The total weight of one servo unit was \SI{79}{\gram}.
The shaft of each servo motor is connected directly to a torsional spring with the respective stiffness for the shoulder, hip, and knee joint as defined by the stiffness matrix $\K$. For each joint, the servo motors were placed off-axis, driving the joint via synchronous belts (Fig.~\ref{fig:hardware}C).

\paragraph*{Communication and power management}
The \ebert robot was equipped with an AAEON UP Squared onboard computer using the Intel Atom x86 platform as the real-time computer. Thus, the system can run fully standalone without any external processing sources.
The 12 servo motors communicated via EtherCAT, and multiple Arduino Micro boards were utilized for communication with additional sensors. This included the link-side hall sensors to read position values after the spring, which were identical to the ones used in the servo modules, as well as an inertial measurement unit (Bosch BNO055) on the base body. The communication interface between the Arduino boards and the control host was implemented via isochronous USB~2.0 to allow for synchronous data exchange of \SI{1}{\kilo \hertz}.
The robot utilizes \SI{12}{\volt} input voltages and can be powered by an external power supply or exchangeable LiPo batteries. A battery pack with \SI{111}{\watt \hour} capacity allows an operation time of up to two hours.

\paragraph*{High-level control}
The high-level controllers were implemented in MATLAB Simulink, which provided commands to all 12 motors simultaneously. The command could be encoded as motor position or velocity control, as well as current control, which acted analogously to torque control, giving the correct scaling. The control loop ran at \SI{1}{\kilo \hertz}.
The information of all sensors and encoders incorporated in the robotic hardware was read out and post-processed in the Simulink model, and updated motor commands were sent back to the motors. As middleware, the DLR-developed \textit{Links and Nodes} was used \cite{schmidt2026links}.
A Python wrapper allows tuning control parameters from a Logitech gamepad.

\paragraph*{Dimensions}
The overall dimensions of the final \ebert hardware match the intended dimensions of a small dog \cite{morey1992dogsize} with a height of \SI{30}{\centi \meter} and a length of \SI{33}{\centi \meter}. The total weight of the completed robot matched the initial bioinspired approximations well at \SI{4.56}{\kilo\gram}. However, when adding the battery of \SI{0.65}{\kilo \gram} the completed system was slightly heavier at \SI{5.2}{\kilo\gram}.
All final dimensions of \ebert are summarized in Fig.~\ref{fig:hardware}A and Tab.~\ref{tab:spec}.

\begin{table}[t]
    \centering
    \caption{\textbf{Overview of all relevant parameters of \ebert.} \\}
    \begin{tabular}{rr|rr|c}
    \textbf{Parameter} & \textbf{Value} & \textbf{Parameter} & \textbf{Value} & \textbf{Unit}  \\
    overall length & 462 & overall width & 265 & [\si{\milli \meter}] \\
    maximum height & 296 & minimum height & 145 & [\si{\milli \meter}] \\
    stance length & 330 & stance width  & 180 & [\si{\milli \meter}] \\
    leg segment length & 120 &  & & [\si{\milli \meter}] \\
    total weight & 4568 %
    & battery weight & 655 & [\si{\gram}] \\
    shoulder stiffness & 13 & hip/knee stiffness & 6.55 & [\si{\newton \meter \per \radian}] \\
    motor max speed & 12 & & & [\si{\radian \per \second}] \\
    motor max torque & 3 & & & [\si{\newton \meter}]\\
    joint max torque & 5 & & & [\si{\newton \meter}]\\
    control rate & 1 & & & [\si{\kilo \hertz}]
    \end{tabular}
    \label{tab:spec}
\end{table}

\subsection{Determining Initial Configuration for Locomotion}
\addcontentsline{toc}{subsection}{Determining Initial Configuration for Locomotion}
\phantomsection
\label{sec:app:grid}
As outlined in the Results section \textit{\nameref{sec:results:locomodes}}, for a first intuition about a suitable initial joint configuration to develop locomotion, we once more turned to biology. Due to the similarity in size and weight, especially dogs appeared as a promising biological inspiration for \ebert. As mentioned, the Center of Mass (COM) in dogs is usually shifted to the front \cite{Michael2024dogCOM}, inherently aiding the body to \textit{fall} forward, while slightly externally rotated feet enhance stability \cite{Zink2020dogposture}.

\begin{figure*}
    \centering
    \includegraphics[width=\textwidth]{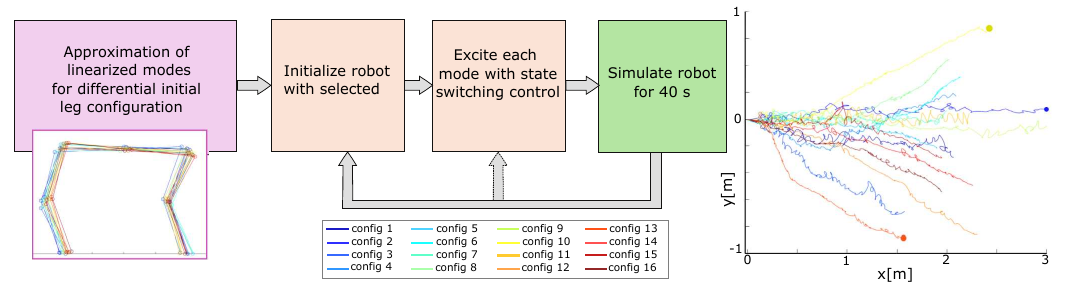}
    \caption{\textbf{Grid search to determine favorable body posture.} 16 joint configurations are investigated, where the hip and shoulder joint angles vary. For each robot configuration, the bang-bang controller is applied to excite the different modes in the realistic robot simulation (with friction), where the feet are not constrained to the ground. After \SI{40}{s} simulation, the traveled distance of the robot without further control is evaluated.
    }
    \label{fig:app:grid_method}
\end{figure*}

To find an adequately tilted body position that is favorable for developing the excited mode oscillations into a useful gait, a grid search was carried out. As visualized in Figure~\ref{fig:app:grid_method}, different robot configurations defined by the commanded hip and knee angles of the front and hind legs were considered. Since a forward tilt is intended, the front leg joints are assumed to be more flexed, while the hind leg joints should be more extended, assuming symmetry about the sagittal plane. \\

Starting from the originally chosen equilibrium positions from Equation~\eqref{eq:leg_angle_def}, the grid search tests all possible combinations of the following angle modifications
\begin{equation}
\begin{split}
    \theta_{F}(\mathrm{hip}) &=  \quad 0.4 +  0.1i \\
    \theta_{F}(\mathrm{knee}) &= -0.4 - 0.1j \\
    \theta_{H}(\mathrm{hip}) &=  \quad 0.4 -  0.1k \\
    \theta_{H}(\mathrm{knee}) &= -0.4 + 0.1l
\end{split}
\ ,
\end{equation}
where $i,j,k,l$ index all combinations of these angle modifications that are tested. Combining the four modifications applied to the front and hind leg angles leads to 16 different body configurations, all realizing a slight forward tilt. The shoulder angles of the front and hind legs remained identical throughout the search.

\begin{figure*}
    \centering
    \includegraphics[width=\textwidth]{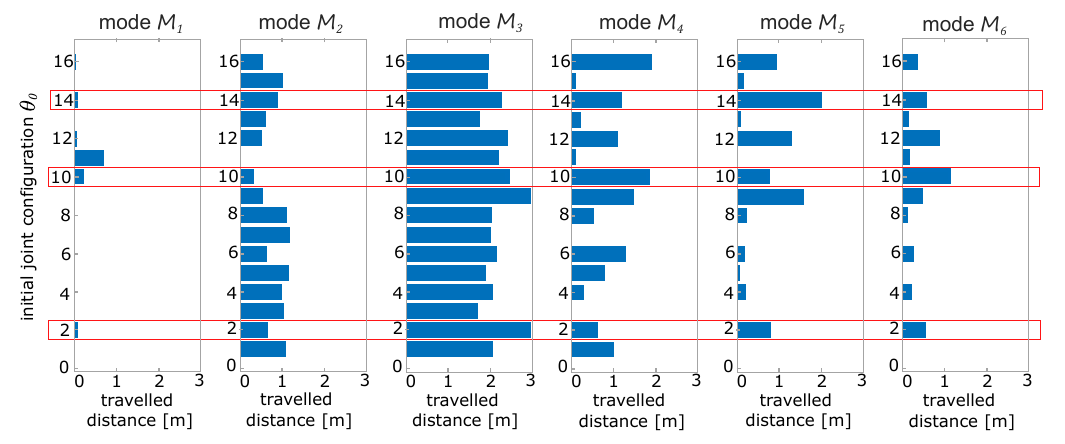}
    \caption{\textbf{Travelled distance for tested initial joint configurations in the different modes.} The configurations 2, 10, and 14 achieved a forward motion with all six modes.
    }
    \label{fig:app:grid_eval}
\end{figure*}

To test the suitability of the different robot configurations in the different modes, the robot was initialized in the Gazebo simulation of \ebert (including friction) with each considered joint configuration (Fig.~\ref{fig:app:grid_method}). Although the derived linearized eigenvectors assumed all feet to remain in contact with the ground, this constraint was purposefully not realized in the simulation. Instead, the state switching controller from Equation~\eqref{eq:newbang-trigger} was applied with a control signal that was empirically tuned in the beginning to inject enough energy that the robot starts to lift its feet. Solely based on the forward tilt of the body, the robot started to move forward. The simulation was continued for \SI{40}{\second}. Afterward, the distance that the robot traveled with each initial configuration, only applying the state switching controller, was evaluated for every mode separately. The resulting plot showing the xy-plane in which the robots move forward is exemplary, as shown in Figure~\ref{fig:app:grid_eval} on the right.
The traveled distance, defined as the norm distance between the starting point of the robot at the beginning of the simulation and the current location of the robot after \SI{40}{\second} is determined for all modes of each tested joint configuration (Fig.~\ref{fig:app:grid_eval}).

It was intended to identify a favorable initial configuration that leads to a forward motion in all modes as a starting point to develop locomotion gaits when a step is added. For this first exploration, it was decided to consider only a single, generally useful initial configuration for all modes to keep the searchable parameter space small. In further investigations, the best initial configuration for different modes should be considered.
Of all the 16 tested initial joint configurations, only configurations 2, 10, and 14 appeared to enable a forward motion in all modes, as highlighted in red in Figure~\ref{fig:app:grid_eval}.
In the first trials, especially mode $\mathcal{M}_3$ appeared promising to develop into a gait pattern for locomotion, which seemed to be best realizable by configuration~2, being the one stated in Equation~\eqref{eq:loco_config}. Thus, this configuration was chosen for the initial exploration of gaits based on modes, but should be extended in future work to explore other promising leg configurations.

\subsection{Control Parameters for the Mode Excitations}
\addcontentsline{toc}{subsection}{Control Parameters for the Mode Excitations}
\label{sec:app:parameters}
In order to excite the mode oscillations of the damped \ebert system in simulation and hardware as described in Section~\textit{\nameref{sec:results:modevalid} }and \textit{\nameref{sec:results:locomodes}}, the state switching controller based on the work of \cite{lak14} was implemented as described in the method section \textit{\nameref{sec:method:bangbang}}. This controller injected energy into the system scaled by $\hat{\theta}_z$ whenever the robot body naturally came to a halt after arming the trigger when crossing $\epsilon_{tau}$. The parameter values were kept identical if possible, but had to be adapted especially for the faster modes $\mathcal{M}_4-\mathcal{M}_6$ to overcome the minimum friction. For the simulation, a velocity limiter was additionally implemented to mimic the behavior of the motor dynamics in the real system. The same limiter was also added to the hardware to ensure consistency of control architecture and system behavior.

For the initial comparison of the mode excitation and the initial symmetric leg configuration of~Equation~\eqref{eq:leg_angle_def}, the parameters detailed in Table~\ref{tab:append:bert_init} were used to obtain the data shown in Figure~\ref{fig:modes-comp}. \\

\begin{table}
\caption{\textbf{Parameter values of the state controller in \ebert's initial mode comparison.}}
\label{tab:append:bert_init}
\centering
\begin{tabular}{ccccc} %
           &               & \textbf{motor}      & \textbf{bang signal} & \textbf{threshold}    \\
           & \textbf{mode} & \textbf{vel. limit} & $\hat{\theta}_z$  & $\epsilon_{\tau}$ \\
           \hline
\textbf{simulation} 
& $\mathcal{M}_1$    & 1         & 0.05        & 0.05         \\
& $\mathcal{M}_2$    & 1         & 0.05        & 0.05         \\
& $\mathcal{M}_3$    & 1         & 0.05        & 0.05         \\
& $\mathcal{M}_4$    & 1         & 0.1         & 0.05         \\
& $\mathcal{M}_5$    & 1         & 0.1         & 0.05         \\
& $\mathcal{M}_6$    & 1         & 0.1         & 0.05         \\
\hline
\textbf{hardware}   
& $\mathcal{M}_1$    & 1         & 0.05        & 0.05         \\
& $\mathcal{M}_2$    & 1         & 0.05        & 0.05         \\
& $\mathcal{M}_3$    & 1         & 0.05        & 0.05         \\
& $\mathcal{M}_4$    & 1         & 0.1         & 0.0          \\
& $\mathcal{M}_5$    & 1         & 0.05        & 0.0          \\
& $\mathcal{M}_6$    & 1         & 0.05        & 0.0
\end{tabular}
\end{table}

In the later experiments to develop locomotion as detailed in Section~\textit{\nameref{sec:method:steps}}, steps were added to each mode oscillation, where the step length and other parameter scaling were determined through black-box optimization. Here, in addition to the aforementioned parameters, the limit of the motor velocity could be slightly increased. The maximum value for this parameter was \SI{5}{\radian \per \second}, which was determined by the actual hardware capabilities. The determined parameters for each mode $\mathcal{M}_1 - \mathcal{M}_6$ to develop the locomotion pattern shown in Figure~\ref{fig:loco-modes} are presented in Table~\ref{tab:append:bert_loco}.

\begin{table*}[b!]
\caption{\textbf{Parameter values of the state controller for \ebert's mode-based locomotion}}
\label{tab:append:bert_loco}
\centering
\begin{tabular}{ccccccccc}
           &               & \textbf{motor}      & \textbf{bang signal} & \textbf{threshold}    & \multicolumn{4}{c}{\textbf{step length}} \\
           & \textbf{mode} & \textbf{vel. limit} & $\hat{\theta}_z$  & $\epsilon_{tau}$  & [FR & FL & HR & HL] \\
           \hline
\textbf{simulation} 
& $\mathcal{M}_1$    & 3& 0.05  & 0.1 &  [-0.25  & -0.30 & -0.07 & -0.13 ]  \\
& $\mathcal{M}_2$    & 2         & 0.2   & 0.1 &  [-0.25  & -0.15 & -0.25 & -0.15 ]  \\
& $\mathcal{M}_3$    & 5         & 0.1   & 0.1 &  [ 0.00  & -0.30 & -0.0  & -0.10  ]  \\
& $\mathcal{M}_4$    & 5         & 0.7   & 0.0 &  [ 0.00  &  0.00 &  0.0  &  0.00  ]  \\
& $\mathcal{M}_5$    & 5         & 0.2   & 0.1 &  [ 0.30  & -0.30 &  0.35 & -0.40  ] \\
& $\mathcal{M}_6$    & 5         & 0.3   & 0.1 &  [-0.20  & -0.30 & -0.2  &  0.15 ]  \\
\hline
\textbf{hardware} 
& $\mathcal{M}_1$  & 3    & 0.03  & 0.1 &  [-0.07 & -0.15 & -0.07 & -0.25 ]  \\
& $\mathcal{M}_2$    & -         & -     & -   &   -     &  -    &  -    &   -     \\
& $\mathcal{M}_3$    & 5         & 0.1   & 0.1 &  [ 0.00 & -0.25 & 0.00   & -0.25 ]   \\
& $\mathcal{M}_4$    & 3.5       & 0.4   & 0.1 &  [ 0.00 &  0.00 & 0.00   &  0.00  ]  \\
& $\mathcal{M}_5$    & -         & -     & -   &   -     &  -    &   -   &   -   \\
& $\mathcal{M}_6$    & 5         & 0.1   & 0.1 &  [ 0.00 & -0.30 & 0.00   & -0.30  ]
    \end{tabular}
\end{table*}

\end{document}